\documentclass[10pt,twocolumn,letterpaper]{article}

\usepackage[pagenumbers]{iccv}      

\usepackage{multirow}
\usepackage{microtype}
\usepackage{pgfplots}\pgfplotsset{compat=1.18}
\usepackage{pgfplotstable}
\usepackage{graphicx}
\usepackage{tabularx}
\usepackage{array}
\usepackage[accsupp]{axessibility}

\newcolumntype{C}{>{\centering\arraybackslash}X}
\definecolor{iccvblue}{rgb}{0.21,0.49,0.74}
\usepackage[pagebackref,breaklinks,colorlinks,allcolors=iccvblue]{hyperref}

\def\paperID{6331} 
\def\confName{ICCV}
\def\confYear{2025}

\title{DAViD: Data-efficient and Accurate Vision Models from Synthetic Data\textsuperscript{\textasteriskcentered}}

\author{Fatemeh Saleh
\qquad 
Sadegh Aliakbarian
\qquad 
Charlie Hewitt
\qquad 
Lohit Petikam
\qquad 
Xiao-Xian
\\
Antonio Criminisi
\qquad 
Thomas J. Cashman
\qquad 
Tadas Baltru\v{s}aitis\\\\
Microsoft, Cambridge, UK
}

\begin{document}

\twocolumn[{%
\maketitle
   \centering
   \includegraphics[width=1.0\textwidth]{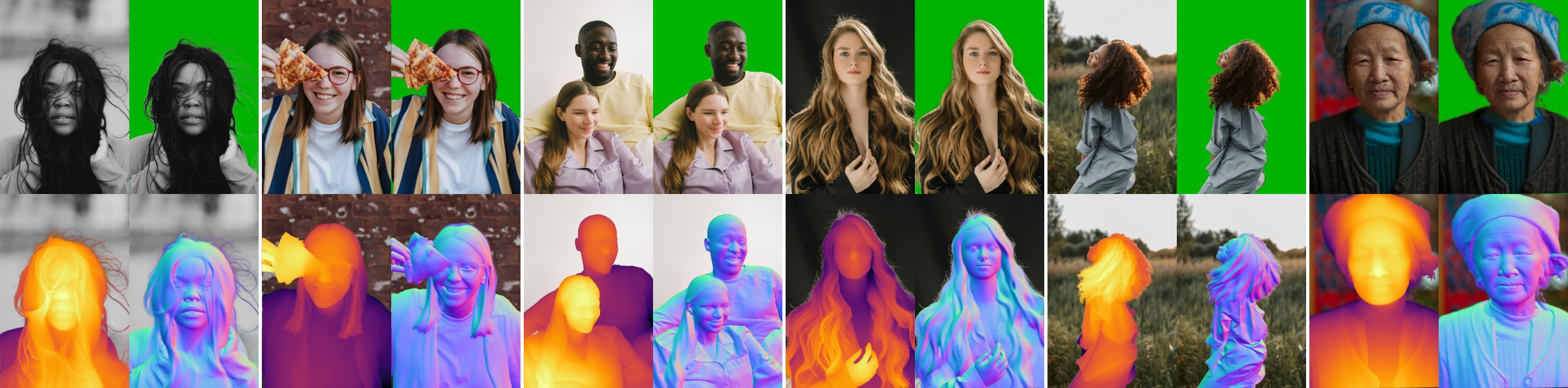}\vspace{-7pt}
   \captionof{figure}{Given a single, real image of a person, our human-centric models, trained entirely on synthetic data, predict accurate relative depth, surface normals, and soft foreground segmentation. 
   Please zoom in to see details such as hair strands, eye glasses and clothes folds.}
   \vspace{1em}
   \label{fig:banner}
}]

\begingroup
\renewcommand\thefootnote{\textasteriskcentered}
\footnotetext{DAViD also references Michelangelo’s David—an iconic symbol of anatomical precision—and the David vs. Goliath story, reflecting our small yet powerful dataset and models.}

\begin{abstract}
The state of the art in human-centric computer vision achieves high accuracy and robustness across a diverse range of tasks.
The most effective models in this domain have billions of parameters, thus requiring extremely large datasets, expensive training regimes, and compute-intensive inference.
In this paper, we demonstrate that it is possible to train models on much smaller but high-fidelity synthetic datasets, with no loss in accuracy and higher efficiency.
Using synthetic \emph{training} data provides us with excellent levels of detail and perfect labels, while providing strong guarantees for data provenance, usage rights, and user consent. 
Procedural data synthesis also provides us with explicit control on data diversity, that we can use to address unfairness in the models we train.
Extensive quantitative assessment on \emph{real} input images demonstrates accuracy of our models on three dense prediction tasks: depth estimation, surface normal estimation, and soft foreground segmentation.
Our models require only a fraction of the cost of training and inference when compared with foundational models of similar accuracy.
Our human-centric synthetic dataset and trained models are available at \url{https://aka.ms/DAViD}.

\end{abstract}

\section{Introduction}
\begin{figure}
    \centering
    \begin{tikzpicture}
  \begin{axis}[
    width=1.\columnwidth,
    height=0.81\columnwidth,
    xlabel={GMACs},
    ylabel={RMSE},
    tick label style={font=\footnotesize},
    label style={font=\footnotesize},
    title style={font=\footnotesize},
    scatter src=explicit,
    point meta=explicit,
    xmin=0,
    xmax=10000,
    ymin=0.1,
    ymax=0.5,
    grid=both,
    legend image post style={mark size=2pt}
  ]
  
  \addplot[
    only marks,
    mark=*,
    color=gray,
    mark options={draw=gray, fill=gray},
    mark size=2163*0.012 pt,
  ] table [x=GMACs, y=RMSE, meta=Params] {
GMACs    RMSE     Params
8709     0.35    2163
  };\node at (axis cs:9500,0.41) [anchor=south east, black] {\scriptsize DepthPro};
    \addplot[
    only marks,
    mark=x,
    color=orange,
    mark options={draw=white, fill=orange},
  ] table [x=GMACs, y=RMSE, meta=Params] {
GMACs    RMSE     Params
8709     0.35    2163
  };

  \addplot[
    only marks,
    mark=*,
    mark options={draw=teal, fill=teal},
    mark size=335*0.012 pt,
  ] table [x=GMACs, y=RMSE, meta=Params] {
GMACs    RMSE     Params
587      0.433    335
  };
  \node at (axis cs:660,0.413) [anchor=south west, black] {\scriptsize DepthAnythingV2-Large};
    \addplot[
    only marks,
    mark=x,
    mark options={draw=white, fill=blue},
  ] table [x=GMACs, y=RMSE, meta=Params] {
GMACs    RMSE     Params
587      0.433    335
  };
  \addplot[
    only marks,
    mark=*,
    mark options={draw=teal, fill=teal},
    mark size=90*0.012 pt,
  ] table [x=GMACs, y=RMSE, meta=Params] {
GMACs    RMSE     Params
160      0.445    90
  };

  \node at (axis cs:160,0.445) [anchor=south west, black] {\scriptsize DepthAnythingV2-Base};

  \addplot[
    only marks,
    mark=*,
    mark options={draw=olive, fill=olive},
    mark size=336*0.012 pt,
  ] table [x=GMACs, y=RMSE, meta=Params] {
GMACs    RMSE     Params
621      0.312    336
  };
  \node at (axis cs:730,0.31) [anchor=south west, black] {\scriptsize Sapiens-0.3B};
  \addplot[
    only marks,
    mark=x,
    mark options={draw=white, fill=red},
  ] table [x=GMACs, y=RMSE, meta=Params] {
GMACs    RMSE     Params
621      0.312    336
  };
  \addplot[
    only marks,
    mark=*,
    mark options={draw=olive, fill=olive},
    mark size=2163*0.012 pt,
  ] table [x=GMACs, y=RMSE, meta=Params] {
GMACs    RMSE     Params
4354.5      0.170    2163
  };
  \node at (axis cs:4354,0.230) [anchor=south, black] {\scriptsize Sapiens-2B};
    \addplot[
    only marks,
    mark=x,
    mark options={draw=white, fill=red},
  ] table [x=GMACs, y=RMSE, meta=Params] {
GMACs    RMSE     Params
4354.5      0.170    2163
  };
  \addplot[
    only marks,
    mark=*,
    color=purple,
    mark options={draw=purple, fill=purple},
    mark size=114*0.012 pt,
  ] table [x=GMACs, y=RMSE, meta=Params] {
GMACs    RMSE     Params
154      0.21    114
  };
  \node at (axis cs:154,0.207) [anchor=south west, black] {\scriptsize Ours-Base};
  
  \addplot[
    only marks,
    mark=*,
    color=purple,
    mark options={draw=purple, fill=purple},
    mark size=347*0.012 pt,
  ] table [x=GMACs, y=RMSE, meta=Params] {
GMACs    RMSE     Params
268      0.19    347
  };
  \node at (axis cs:268,0.18) [anchor=south west, black] {\scriptsize Ours-Large};
    \addplot[
    only marks,
    mark=x,
    color=black,
    mark options={draw=white, fill=black},
  ] table [x=GMACs, y=RMSE, meta=Params] {
GMACs    RMSE     Params
268      0.19    347
  };


  \end{axis}
\end{tikzpicture}
    \caption{Compute cost vs error, comparing our method with state-of-the-art depth estimation models. Compute cost and error are measured with giga-multiply-accumulate count (GMACs) and root-mean-squared error (RMSE), respectively, on the combination of Goliath~\cite{martinez2024codec} and Hi4D~\cite{yin2023hi4d} datasets. The radius of each marker is proportional to the number of model parameters. The most efficient and accurate models are in the lower-left corner. 
}
    \label{fig:perf_acc}
\end{figure}
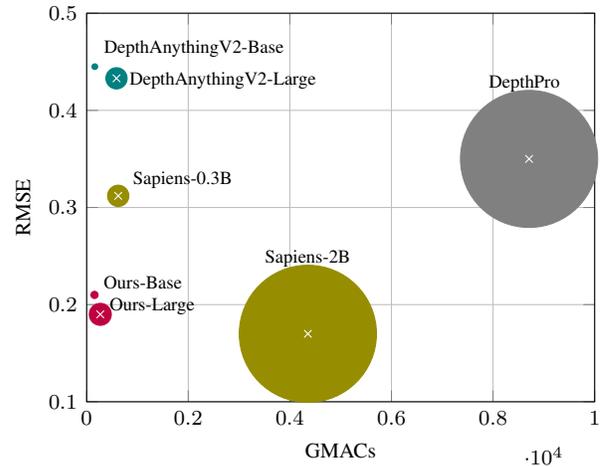
Progress in human-centric computer vision has been driven in large part by advances in data. This is both due to the scale and diversity of available data \cite{lin2014microsoft, Yang2015WIDERFA, khirodkar2024_sapiens, karkkainenfairface} and the quality of annotations \cite{jin2020whole, SAGONAS20163}. 
Some types of ground-truth labels can be annotated by humans (e.g., landmarks \cite{jin2020whole, SAGONAS20163}, coarse semantic classes \cite{wang2019face, liu2020new} and bounding boxes \cite{Yang2015WIDERFA}).
However, labels such as per-pixel depth, normals, or dense landmarks are significantly more challenging or even impossible for humans to annotate.
Gathering such annotations 
often relies on complex camera rigs \cite{martinez2024codec, Soubhik2019, boyne2022find,BrostowPAMI11,lightstage2007}, or specialist sensors \cite{KinectFaceDB}.
This leads to imperfect ground truth annotations, as they are derived from photogrammetry or noisy sensors.
Further, in-lab captures significantly limit diversity of subjects and environments, as it is extremely challenging to capture truly in-the-wild data for such tasks.
Training only on such datasets leads to models that produce coarse or inaccurate predictions, and which struggle to generalize outside the domain of the collected data \cite{Black_CVPR_2023,khirodkar2024_sapiens, yang2023synbody}.

In order to satisfy requirements for scale, diversity and high fidelity of annotations, recent approaches rely on large quantities of diverse data and a smaller amount of annotated data \cite{khirodkar2024_sapiens, yang2023synbody, bochkovskii2024depth}. 
These techniques typically follow a two-stage approach: first, large-scale pretraining on real data with no or lower-quality ground truth, followed by fine-tuning on data with high-quality ground-truth annotations.
These methods show good accuracy, but come at a considerable computational model training cost, and require complex multi-stage training. 
Finally, the accuracy of such methods is limited by the quality of the data used for fine-tuning.
For example, Sapiens~\cite{khirodkar2024_sapiens} relies on coarse synthetic data, and struggles to capture fine details such as facial wrinkles, eyelids, or subtle texture variations in clothing (see \cref{fig:sx_gt_comparison} for ground truth quality and \cref{fig:depth_qual} for qualitative results).

Instead, we propose to tackle both diversity and fidelity of training data through the use of procedurally-generated synthetic data \cite{hewitt2024look}.
We demonstrate that a single high-fidelity dataset is sufficient to tackle multiple dense prediction tasks and achieve state-of-the-art accuracy.
Our approach requires a fraction of the data size, model size, computational complexity, and training time of competing approaches, all without sacrificing model accuracy on challenging cross-dataset evaluations (see Fig.~\ref{fig:perf_acc}).
We demonstrate this on three challenging dense prediction tasks: relative depth estimation, surface normal estimation, and soft foreground segmentation, with our models capturing subtle details, handling thin structures, and maintaining accurate human proportions.

Our approach is different from techniques such as DepthPro~\cite{bochkovskii2024depth}, DepthAnything-v2~\cite{yang2025depth}, and Sapiens~\cite{khirodkar2024_sapiens}, which either develop large, task-specific models, employ complex training regimes, or rely on large-scale data collections.
We use a single architecture and a single dataset to tackle all three tasks.
Importantly, training on synthetic data alone allows us to verify compliance with privacy, copyright, licensing, consent and diversity requirements, which would be more challenging to achieve with large datasets of real images.
The core contribution of this paper is to demonstrate a fundamentally more \emph{efficient} paradigm for human-centric vision. Our work demonstrates that it is possible to train performant and state-of-the-art human-centric models in a fraction of the time and on a fraction of data by relying solely on high-quality synthetic data.
Details of how to access the SynthHuman dataset and trained models are available on the project website: \url{https://aka.ms/DAViD}.

\section{Related Work}

\noindent\textbf{Human vision data.}
The availability of high-quality training data has boosted accuracy of recent computer vision models~\cite{DenDon09Imagenet,bochkovskii2024depth, Radford2021clip, oquab2024dinov}, with no exception for human-centric tasks \cite{Yang2015WIDERFA,Black_CVPR_2023,liu2020new}.
This is especially true for face detection \cite{Yang2015WIDERFA}, pose estimation \cite{Black_CVPR_2023}, landmark localization \cite{SAGONAS20163}, and semantic segmentation \cite{liu2020new}, where manual annotation is feasible with current tools and methodologies~\cite{jin2020whole, lin2014microsoft,sagonas2016300}.
However, obtaining pixel-wise annotations manually for tasks such as matting, depth and surface normals is much harder~\cite{bochkovskii2024depth, yang2025depth,Cordts2016Cityscapes,ke2022modnet}. 
To alleviate this, some approaches have relied on curated multi-view real-image datasets  to reconstruct human meshes~\cite{martinez2024codec,boyne2022find, yin2023hi4d}.
While providing rich annotations, these datasets are limited in subject and environment diversity, due to the high costs of data collection.
Further, as they rely on model-fitting or photogrammetry, they struggle with very thin structures like hair, reflective or semi-transparent surfaces like glasses and eyes, and are not able to capture high-frequency details (see \cref{fig:sx_gt_comparison}).
Our procedural synthetic data generation pipeline allows us to create data that is both diverse and has pixel-perfect labels.

\noindent\textbf{Synthetic training datasets.}
Synthetic data
has emerged as an alternative to overcome the annotation bottleneck in human-centric vision tasks. 
Early efforts focused on rendering pre-defined 3D human meshes acquired through photogrammetry \cite{Patel2021agora, han2023high, tao2021function4d}. 
While allowing for automatic dense annotations, the resulting data is limited by lack of reflective objects (e.g., glasses) and the quality of meshes, which are often low-fidelity, especially around hair, eyes, and digits.
Procedural synthetic data can provide improved fidelity and diversity.
For example, \citet{wood2021fake} demonstrated how a procedural synthetic data pipeline can be used to train facial landmark detection and face parsing models. 
BEDLAM \cite{Black_CVPR_2023} offers a full-body synthetic pipeline, featuring clothed subjects captured in diverse lighting environments.
Built on the SMPL-X body model~\cite{SMPL-X:2019}, BEDLAM introduces variability in body shape and pose, however it lacks high-fidelity faces, hair, and mesh-based environments.
Our work builds upon the synthetic data pipeline of \citet{hewitt2024look}, and allows for high-fidelity expressive bodies and faces.
Further, it benefits from artist-created accessories, clothing, and environments to increase the diversity of generated data.
This allows the models trained on our dataset to exhibit high accuracy and to better generalize to unseen scenarios. 

\noindent\textbf{Training on Synthetic Data.}
To address data diversity and quality issues, hybrid data strategies have been proposed. 
Depth Anything v2~\cite{yang2025depth}, uses a robust teacher model (DINOv2-G) which is trained exclusively on 595K synthetic images. This model then generates precise pseudo ground truth for a large collection of 62M unlabeled real images, which are subsequently used to train a student model.
DepthPro~\cite{bochkovskii2024depth} follows a two-stage training curriculum. In the first stage, the model is trained on a mix of multiple real datasets with noisy ground truth, utilizing carefully selected loss functions to improve convergence. In the second stage, the model is trained on synthetic datasets with perfect ground truth.

More recently, Sapiens~\cite{khirodkar2024_sapiens} propose pre-training a large model on 300M real images using self-supervised learning and fine-tuning it on 500K high-resolution synthetic images for depth and surface normal estimation.
While achieving promising results, it comes at a significant computational cost. 
Pre-training the largest variant required 18 days on 1,024 A100 GPUs\footnote{The authors did not discuss the computational costs of fine-tuning.}.
In contrast, our work simplifies the training strategy and eliminates the need for data mixing by using a single small-scale and high-fidelity dataset.

\section{Method}

\subsection{SynthHuman: Human-centric Synthetic Data}
\label{sec:sx_data}

To train our models, we use exclusively synthetic data. To this end, a common choice is to use scan-based synthetic data generation ~\cite{renderpeople,tao2021function4d}.
However, their quality is often limited by the 3D scanning technology used and the 3D mesh representation (see \cref{fig:sx_gt_comparison} for the comparison of the ground truth quality). 
Recently, higher fidelity synthetic data, following the practices of games and visual effects, has been demonstrated to be more effective for certain tasks such as landmark prediction and 3D reconstruction~\cite{wood2021fake,hewitt2024look,Black_CVPR_2023}.
In this work, we extend the use of such high-fidelity synthetic data to \emph{dense} prediction tasks where realism and annotation quality are even more critical, and for which annotations on real data are often impossible.
Specifically, we use the data generation pipeline of \citet{hewitt2024look}, incorporating the updated face model of \citet{petikam2024eyelid}, to create a human-centric synthetic dataset with a high degree of realism, as well as high-fidelity ground-truth annotations.
Our \emph{SynthHuman}, dataset contains 300K images of resolution $384 \times 512$, covering examples of faces, upper body, and full body scenarios equally. 
Along with the RGB rendered image, each sample includes soft foreground mask, surface normals, and depth ground-truth annotations, used to train our models. 
We design SynthHuman such that it is diverse in terms of poses, environments, lighting, and appearances, and not tailored to any specific evaluation set.
This allows us to train models that generalize across a range of benchmark datasets, as well as on in-the-wild data.
Examples of our training data are shown in \cref{fig:sx_training_data}.
Rendering the dataset took 72 hours on a cluster of 300 machines with M60 GPUs\footnote{The cost is equivalent to 2 weeks of an A100 machine with 4 GPUs.}.

\begin{figure}
    \centering
    \includegraphics[width=\linewidth]{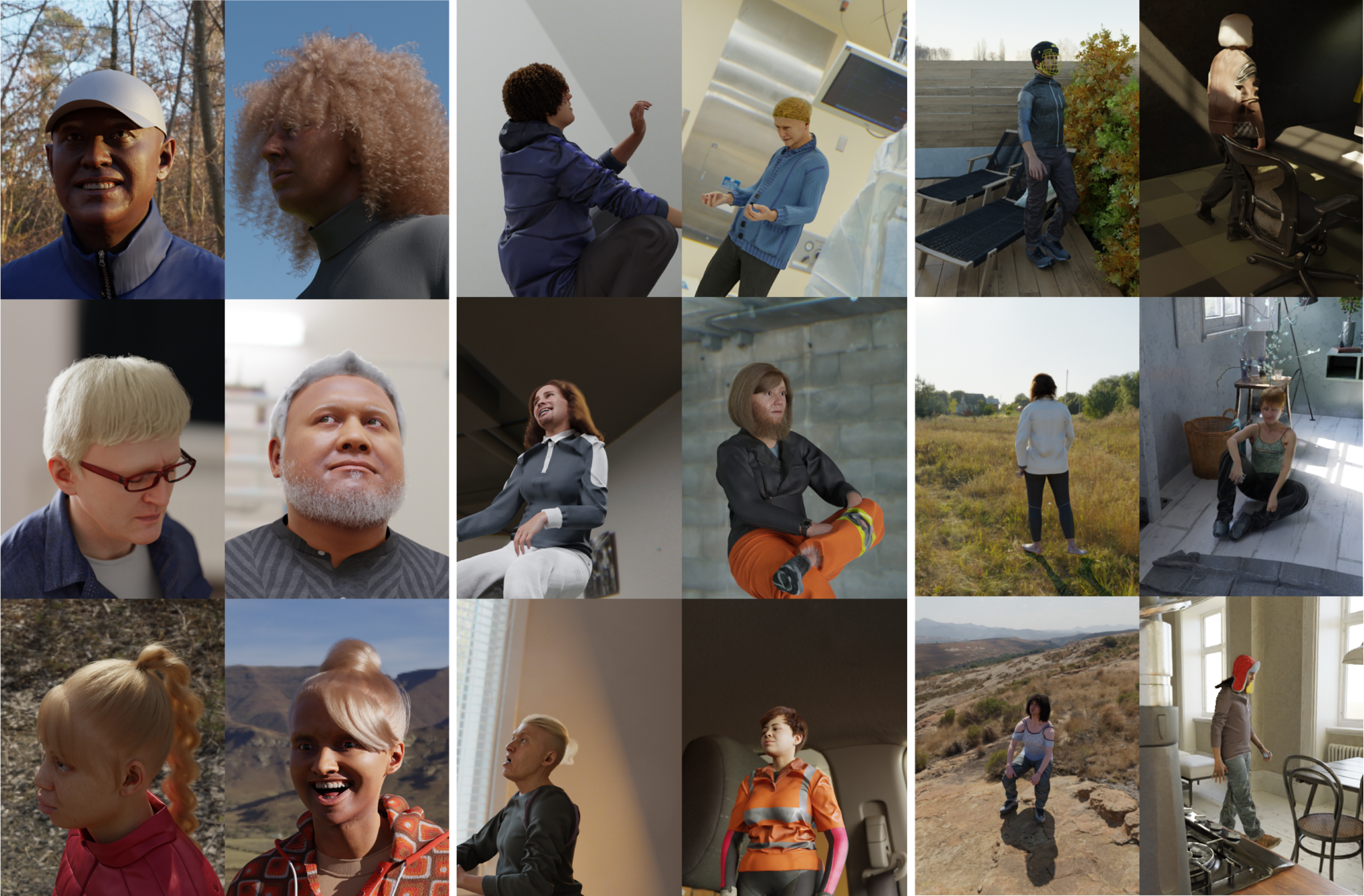}
    \vspace{-1.5em}
    \caption{Random samples of our synthetic training images for the face, upper and fully body.}
    \label{fig:sx_training_data}
\end{figure}

\begin{figure}
    \centering
    \scriptsize
    \begin{tabularx}{\linewidth}{CCC}
         THuman~\cite{tao2021function4d} & Renderpeople~\cite{renderpeople} & SynthHuman \\
    \end{tabularx}
    \includegraphics[width=\linewidth]{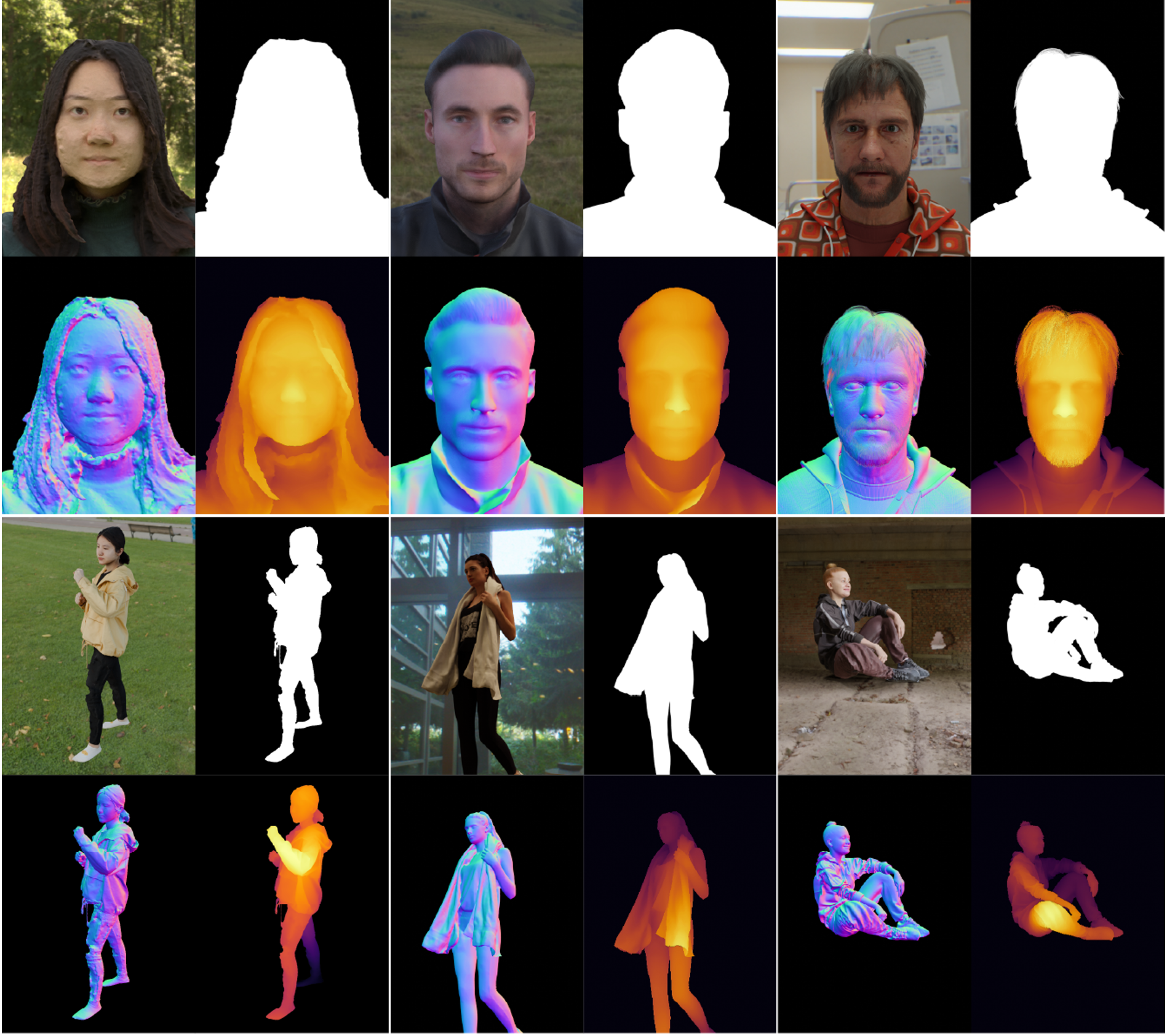}
    \vspace{-1.5em}
    \caption{Ground-truth annotations for depth, surface normals and soft foreground segmentation for our synthetic data in comparison to synthetic data used in other work. Note the significantly higher fidelity annotations, particularly for hair and clothing, in our data. Our data is also free of scanning artifacts common in THuman data.}
    \label{fig:sx_gt_comparison}
\end{figure}

Our results demonstrate that using this high-quality data enables very accurate results with smaller models and less data, leading to a far more economical training and inference.


\subsection{Model Architecture}
\label{subsec:architecture}
We use a single model architecture (with varying number of output channels) to tackle the three dense prediction tasks. 
We adapt the dense prediction transformer (DPT)~\cite{ranftl2021vision} to handle variable input resolutions efficiently.
\begin{figure}
    \centering
    \includegraphics[width=\linewidth]{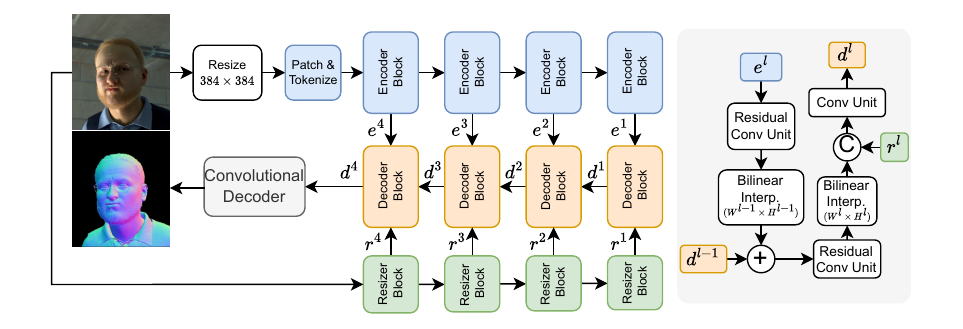}
    \vspace{-1.5em}
    \caption{(Left) Overview of the model architecture, with an example of surface normal prediction. (Right) Our decoder block.}
    \label{fig:method}
\end{figure}
As illustrated in \cref{fig:method}, our architecture has three main components: encoder blocks, resizer blocks, and decoder blocks. 

\noindent\textbf{Encoder.}
We use the ViT~\cite{dosovitskiy2020image} architecture as our image encoder backbone. The encoder design follows DPT's encoder with $Read_{proj}$ as the read operation (see \citet{ranftl2021vision}).
\begin{align}
    e^l = \text{mlp}(\text{cat}(\texttt{CLS}^l, t_i^l))
\end{align}
wherein $e^l$ is the sequence of updated visual tokens at layer $l$ after projection, $\texttt{CLS}^l$ is the optimized \texttt{CLS} token at layer $l$, and $t_i^l$ is the $i^{\text{th}}$ visual token at layer $l$. 

\noindent\textbf{Resizer.}
While we keep a fixed resolution for the input to the ViT encoder (specifically $384\times 384$), we utilize another light-weight fully convolutional image encoder to carry information at any resolution. 
These features are computed on the original image size and are used in the decoder blocks, described below. 
Each image resizer block is a convolutional module, defined as $g$. Particularly, $r^l = g^l(r^{l-1})$ at layer $l$ computes new features at half the resolution of its input tensor.
To form the full resizer module, we stack four resizer blocks, similar to the number of encoder blocks we use to extract intermediate features. Note that this is to alleviate the need for running the ViT encoder on a potentially higher-resolution image, which comes at a much higher computational cost due to the quadratic nature of self-attention.

\noindent\textbf{Decoder.}
The decoder aims at generating feature representations that the convolutional head (described below) can generate the output from. Each decoder block in the decoder module, as depicted in \cref{fig:method} (right), works with 3 inputs: (1) The output from previous decoder block, $d^.$, if available. (2) Corresponding feature from the encoder, $e^.$. (3) Corresponding feature from image resizer, $r^.$. \begin{align}
    d^l_{\textrm{int}} & = \text{RConv}(d^{l-1} + \text{Interp}(\text{RConv}(e^l)))\nonumber \\ 
    d^l & = \text{Conv}([r^l, \text{Interp}(d^l_{int})])
\end{align}
where $d^l_{\textrm{int}}$ is an intermediate feature used for internal computations in the decoder block, Interp is the bilinear interpolation, Conv is the convolutional unit and RConv is the residual convolutional unit. In particular, the decoder block first fuses the output of previous decoder block with the corresponding encoder features by first upsampling a learned residual from the encoder feature and adding it to previously decoded features. The resulting representation is then transformed into another feature map via a residual convolutional unit, followed by upsampling to the resolution of the corresponding image resizer features. The results are then concatenated with the image resizer features, producing the output after going through a convolutional unit.

\noindent\textbf{Convolutional Head.}
The convolutional head for each task also follows the design of DPT, with different number of output channels for different tasks: 1 for portrait matting, 1 for relative depth, and 3 for surface normals. 

\noindent\textbf{Remark on Resizer.}
We explicitly use a fixed-size input to the ViT for constant inference cost of the encoder and handle variable resolutions with the Resizer and the modified decoder (Fig.~\ref{fig:method} (right)).
This is a more efficient alternative to increasing the number of visual tokens (as done in, e.g., Sapiens~\cite{khirodkar2024_sapiens})\footnote{Additionally, it allows us to evaluate any resolution as opposed to constraining on a multiplier of $p$.} if the input image has higher resolution. 
We empirically observed that not only is this faster, it also yields compelling results capturing fine-grained details (see supplementary material for results).

\subsection{Loss Functions}
Having presented the model architecture, 
next we present the training losses used to address our three prediction tasks.

\noindent\textbf{Soft Foreground Segmentation.} 
For this task, the model only predicts a soft alpha mask, $\hat{\alpha}$, without learning the composition. To train the model, we use a loss function as
\begin{equation}
    \mathcal{L}_{\alpha} = \mathcal{L}_{\textrm{BCE}} + \mathcal{L}_{L1} + \mathcal{L}_{\textrm{dice}} + \omega_{\textrm{lap}}\mathcal{L}_{\textrm{lap}}
\end{equation}
wherein $\mathcal{L}_{\textrm{BCE}}$ is the binary cross-entropy loss, 
$\mathcal{L}_{L1}$ is the $L1$ loss, $\mathcal{L}_{\textrm{dice}}$ is the dice loss~\cite{sudre2017generalised}, and finally $\mathcal{L}_{\textrm{lap}}$ is the $L1$ reconstruction loss between the Laplacian pyramid representation~\cite{hou2019context} of the ground truth soft mask, $\alpha$, and $\hat{\alpha}$. All terms except for $\mathcal{L}_{\textrm{lap}}$ are weighted equally. We observed that $\omega_{\textrm{lap}} < 1$ leads to better accuracy.

\noindent\textbf{Surface Normal Estimation.}
The model predicts the per-pixel $xyz$ components of the normal vector, forming a 3-channel output, 
$\hat{\eta}$, at the same resolution as the input image. Our model is trained to maximize the alignment between the predicted normalized and ground truth surface normal maps, $\hat{\eta}$ and $\eta$, respectively, using cosine similarity,
$\mathcal{L}_{\eta} =  1 - \eta . \hat{\eta}$, computed on the foreground region.

\noindent\textbf{Monocular Relative Depth Estimation.}
For relative depth estimation, we first normalize the ground-truth metric depth, $d^*$, by $d=\frac{d^* - \min(d^*)}{\max(d^*)- \min(d^*)}$. The model estimates a relative depth, $\hat{d}$, that closely matches the normalized ground-truth depth. We use a shift-and-scale-invariant loss~\cite{lasinger2019towards}. To encourage sharper boundaries, we supervise gradients of the predictions~\cite{bochkovskii2024depth}
\begin{equation}
    \mathcal{L}_{d} = \mathcal{L}_{\textrm{MSE}}(s\hat{d}+t, d) + \omega_{\textrm{grad}}\mathcal{L}_{\textrm{grad}}(s\hat{d}+t, d)
\end{equation}
where $s$ and $t$ are scale and shift scalars, computed using the method of \citet{lasinger2019towards}. We compute the depth loss on the foreground region only.

\section{Experiments}

\begin{figure}
    \centering
    \footnotesize
    \begin{tabularx}{0.95\linewidth}{CCCCC}
         Input & \multicolumn{2}{c}{Ours} & \multicolumn{2}{c}{Sapiens-2B~\cite{khirodkar2024_sapiens}} \\
         image & \scriptsize Normals & \scriptsize Depth & \scriptsize Normals & \scriptsize Depth \\
    \end{tabularx}
    \includegraphics[width=0.95\linewidth]{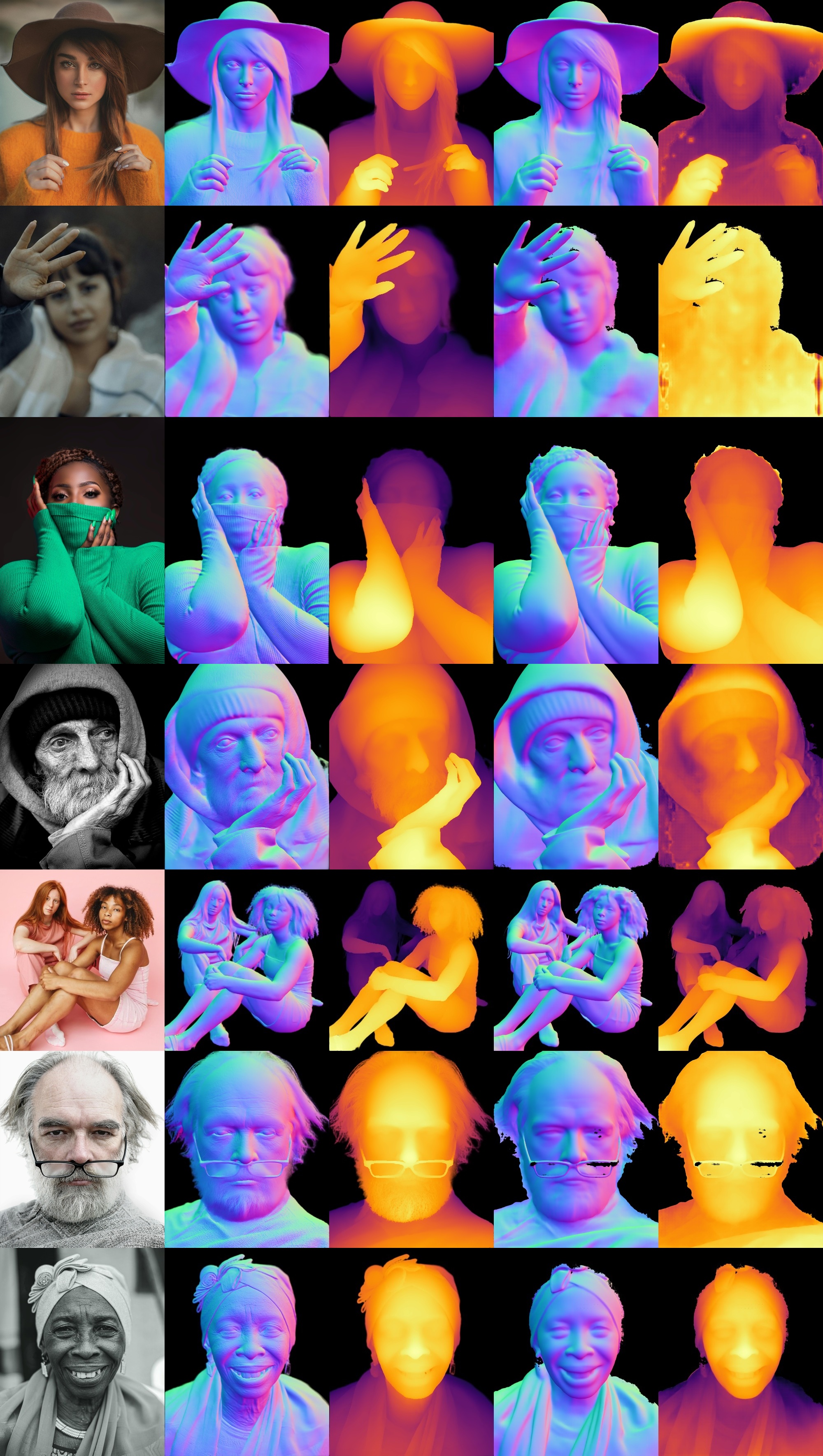}
    \vspace{-0.5em}
    \caption{Qualitative comparison between our method and  Sapiens~\cite{khirodkar2024_sapiens} on challenging in-the-wild images. We use the publicly released Sapiens-1B segmentation model for Sapiens' foreground segmentation. 
    See supplementary material for more results.
    }
    \label{fig:depth_qual}
\end{figure}

\subsection{Implementation details}
As mentioned in \cref{sec:sx_data}, we train our models on 300,000 Synthetic images from our SynthHuman dataset. For each task, we train the model for 100 epochs, with an AdamW optimizer~\cite{loshchilov2017decoupled} with a starting learning rate of $1e^{-5}$ decreasing following a cosine annealing scheduler~\cite{loshchilov2016sgdr}. We use a batch size of 24 on each GPU of a A100$\times$4 compute node.
The training images are rendered at the resolution of 384$\times$512, with an aspect ratio of 3:4, which aligns with human-centric test sets. Since the original DPT encoder requires a square image, we pad around the image (by replicating the sides) to 512$\times$512. 
We provide more details about augmentation during training in the supplementary materials.
\subsection{Evaluation protocol}
\noindent\textbf{Evaluation Datasets.}
We evaluate our approach on multiple challenging real benchmark datasets. 
We use the Goliath~\cite{martinez2024codec} and Hi4D~\cite{yin2023hi4d} datasets to evaluate our depth and surface normal estimation models\footnote{See supplementary material for additional experiments on the synthetically generated dataset from THuman2.1~\cite{tao2021function4d}, following Sapiens~\cite{khirodkar2024_sapiens}.}. 
Goliath contains data from four subjects, for which we use the head and fully-clothed captures to create three subsets of face, upper body, and full body.
Each subset uses 12 cameras and 16 frames (minus two missing cameras) resulting in total of 2,272 test samples (generation details in the supplementary material).
The Hi4D dataset provides captures of subject pairs interacting. 
Following the evaluation protocol in Sapiens~\cite{khirodkar2024_sapiens}, we selected the same sequences from pairs 28, 32, and 37, which include 6 unique subjects recorded by camera 4. 
This selection results in total of 1,195 multi-human real images for testing.
For soft foreground segmentation, we report our results on the PhotoMatte85~\cite{lin2021real} and PPM-100~\cite{ke2022modnet} datasets. We also provide more results on the P3M~\cite{li2021privacy} dataset in the supplementary materials.

\noindent\textbf{Evaluation Metrics.}
To evaluate depth estimation models, we report the mean absolute value of the relative depth (AbsRel) and the root mean square error (RMSE), following standard practice~\cite{ranftl2020towards,khirodkar2024_sapiens,yang2025depth}. 
To evaluate surface normal estimation models, we report the standard metrics~\cite{khirodkar2024_sapiens, eftekhar2021omnidata} of mean and median angular error, as well as the percentage of pixels within $t^\circ$ error for $t\in\{11.25, 22.5, 30\}$. 
For soft foreground segmentation, we report common metrics following \citet{li2021privacy} including sum of absolute differences (SAD), mean squared error (MSE), mean absolute difference (MAD), and Connectivity (Conn.).

\subsection{Comparison to the state of the art}
Unlike prior approaches, which rely on distinct models and/or datasets for each task and often require task-specific tuning, architectural modifications, or additional processing modules (e.g., refiner networks or guided filters for matting, multi-resolution decoders or focal length prediction for depth estimation), our method employs a unified architecture trained on a single dataset. The only variations are the loss functions and the number of output channels. Despite its significantly smaller model size, our approach achieves competitive performance across multiple benchmarks, underscoring the crucial role of high-fidelity training data.

\noindent\textbf{Relative Depth Estimation.}
We evaluate our approach on two challenging real datasets Goliath and Hi4D. \cref{tab:depth_eval} compares our method with the existing state-of-the-art approaches. The variants of our method (Base and Large) demonstrate remarkable performance for such comparably efficient models. 
Our Large variant, with 0.3B parameters, performs on-par with foundation model of Sapiens~\cite{khirodkar2024_sapiens} which has over 2B parameters, while running $\sim$16x faster (measured in MACs).
Our model also outperforms DepthPro~\cite{yang2025depth}, which is trained specifically for sharp depth estimation from high resolution images. 
This shows the quality of the training data plays a major role in the accuracy of the model, and enables training of very simple models without custom designs\footnote{We acknowledge that DepthPro is trained on generic datasets, including human-centric ones, e.g., Bedlam~\cite{Black_CVPR_2023}. Similarly, DepthAnythingV2 and MiDaS are also trained for depth estimation in any scenario.}. 
Not only are our models accurate, they are much smaller and faster than the competing approaches, running at $\sim$48 FPS on NVIDIA A100. 
On the Hi4D dataset, we noticed that our replication of existing baselines lead to negligibly different (better) accuracy for depth estimation task\footnote{We suspect this is due to differences in ground truth rendering.}, so we re-evaluated all methods in \cref{tab:depth_eval} for a fair comparison.
\cref{fig:depth_qual} illustrates the robustness of our approach when tested on the 
long tail (i.e., the less common, more specialized cases) of the distribution of human-centric images.

\begin{table*}
    \centering
    \footnotesize
    \caption{Depth estimation on Goliath and Hi4D dataset.} 
\resizebox{\linewidth}{!}{%
\begin{tabular}{lcccccccccccc}
\toprule
\multirow{2}{*}{Method} & 
\multirow{2}{*}{GFLOPS} & 
\multirow{2}{*}{Params} & 
\multicolumn{2}{c}{Goliath-Face} & 
\multicolumn{2}{c}{Goliath-UpperBody} & 
\multicolumn{2}{c}{Goliath-FullBody} & 
\multicolumn{2}{c}{Hi4D} &
\multicolumn{2}{c}{Averaged over all}
\\
\cmidrule(lr){4-5} \cmidrule(lr){6-7} \cmidrule(lr){8-9} \cmidrule(lr){10-11} \cmidrule(lr){12-13}
&&& RMSE $\downarrow$ & AbsRel $\downarrow$ & RMSE $\downarrow$ & AbsRel $\downarrow$ & RMSE $\downarrow$ & AbsRel $\downarrow$ & RMSE $\downarrow$ & AbsRel $\downarrow$ & RMSE $\downarrow$ & AbsRel $\downarrow$\\
\midrule
MiDaS-DPT\_L~\cite{ranftl2020towards} & - & 0.34B & 0.224 & 0.016  & 0.553 & 0.015 & 0.973 & 0.027  & 0.148 & 0.042 & 0.437 & 0.027 \\
 DepthAnythingV2-L~\cite{yang2025depth} & 1827 & 0.34B & 0.229 & 0.017& 0.492 & 0.014 & 1.039 & 0.029 & 0.130 & 0.034 & 0.433 & 0.025 \\
Sapiens-0.3B~\cite{khirodkar2024_sapiens} & 1242 & 0.34B & 0.179 & \underline{0.012} & 0.368 & 0.010 & 0.690 & 0.019 & 0.116 & 0.035 & 0.312 & 0.021\\
 Sapiens-2B~\cite{khirodkar2024_sapiens} & 8709 & 2.16B & 0.158 & \textbf{0.009} & \textbf{0.204} & \textbf{0.005} & \textbf{0.266} & \textbf{0.007} &  0.095 & 0.030 & \textbf{0.170} & 0.015 \\
Depth-Pro~\cite{bochkovskii2024depth} & 4370 & 0.50B & 0.295 & 0.020 & 0.442 & 0.010 & 0.723 & 0.016 & \underline{0.084} & \textbf{0.018} & 0.350 & 0.016\\
\midrule
Ours-Base & 344 & 0.12B  & \underline{0.142} & \textbf{0.009} & 0.316 & 0.009 & 0.376 & 0.010 & 0.085 & 0.024 & 0.212 & \underline{0.014} \\
Ours-Large & 663 & 0.34B & \textbf{0.140} & \textbf{0.009} & \underline{0.283} & \underline{0.008} & \underline{0.334} & \underline{0.009} & \textbf{0.072} & \underline{0.019} & \underline{0.191} & \textbf{0.012} \\

\bottomrule
\end{tabular}%
}
    \label{tab:depth_eval}
\end{table*}

\noindent\textbf{Surface Normal Estimation.}
We compare our surface normal prediction model on Goliath and Hi4D and report the results in \cref{tab:normal_eval_all}.
Similar to our depth model, our surface normal prediction model achieves very competitive performance while requiring far fewer parameters. Specifically, our model outperforms baselines of similar size, e.g., Sapiens-0.3B and performs on par with largest models, e.g., Sapiens-2B. 
Although Goliath and Hi4D provide good source of real data for testing, the ground-truth annotations are very noisy. \cref{fig:normal_goliath} demonstrates that a large source of error in the metric is the lack of detail in the ground truth, indicating that we may be observing ceiling effects when evaluating our models.
Specifically, detail is lacking for the mouth interior, wrinkles in clothing, and there are incorrect connected regions (attached fingers or arm-body attachment).
As illustrated, our model captures far more detail such as wrinkles, evident in \cref{fig:normal_goliath} and in challenging in-the-wild data depicted in \cref{fig:depth_qual}.
See supplementary material for further discussion on annotation quality of surface normals in the test data.

\begin{figure}
    \centering
    \scriptsize
    \begin{tabularx}{\linewidth}{CCCC}
         Input Image & Our Prediction & Rendered GT & Error Map \\
    \end{tabularx}
    \includegraphics[width=\linewidth]{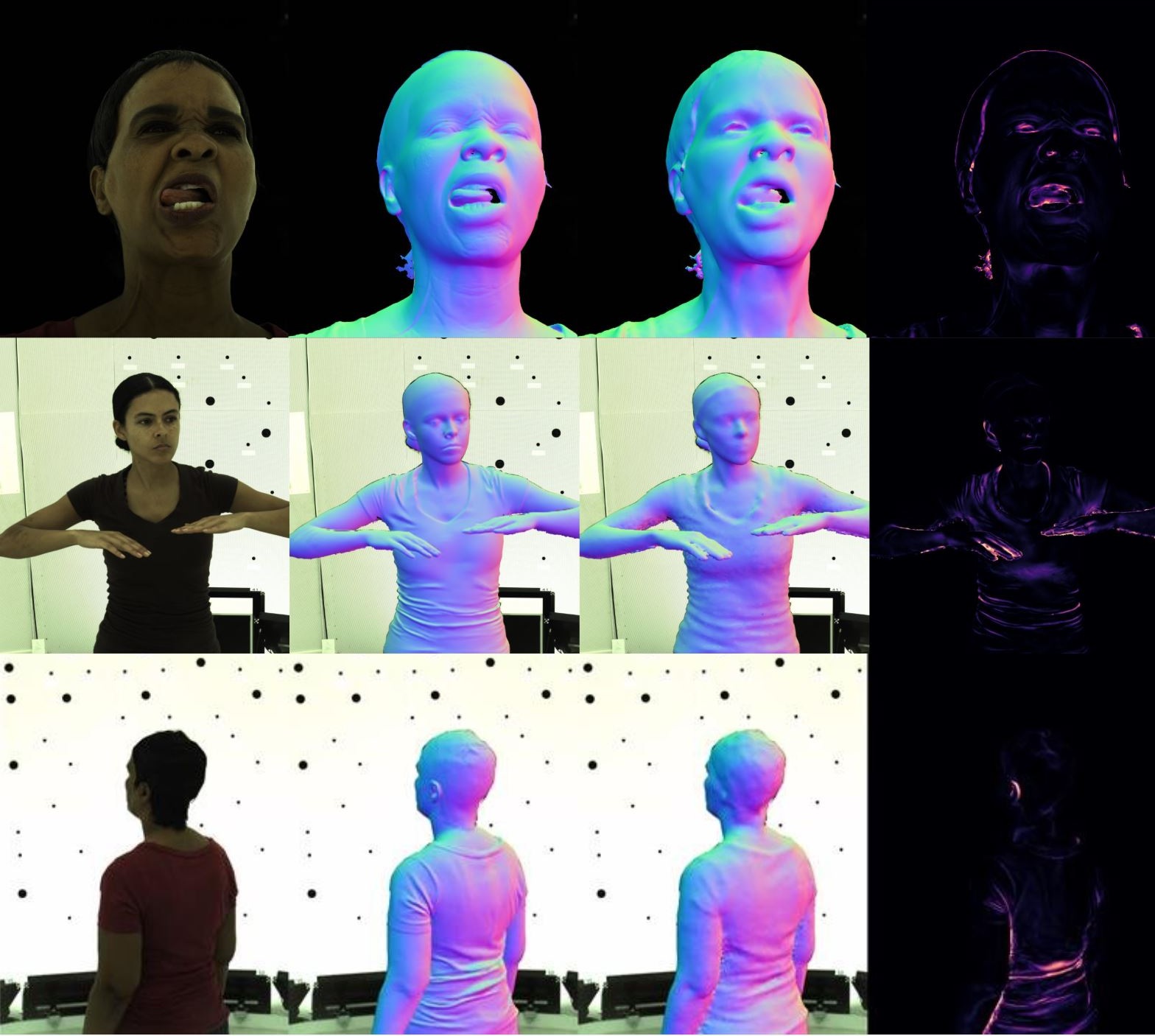}
    \vspace{-1.5em}
    \caption{Qualitative results on Goliath dataset. As shown in the last column (the error map between our prediction and the ground truth), the main source of error is in high frequency details. While our approach captures very fine details, we observed that the ground truth is very coarse, lacking fine-grained details, such as details of mouth interior, face wrinkles as a result of expression, detailed wrinkles in clothing, and separation of body-arm and fingers.}
    \label{fig:normal_goliath}
\end{figure}

\begin{table*}
\centering
\scriptsize
\caption{Surface normal estimation results on Goliath and Hi4D. All results on the Hi4D dataset are taken from~\cite{khirodkar2024_sapiens}.}
\resizebox{\linewidth}{!}{%
\begin{tabular}{l cc c cc c cc c cc c}
\toprule
\multirow{3}{*}{Method} & 
\multicolumn{3}{c}{Goliath-Face} & 
\multicolumn{3}{c}{Goliath-UpperBody} &
\multicolumn{3}{c}{Goliath-FullBody} &
\multicolumn{3}{c}{Hi4D} \\
\cmidrule(lr){2-4} \cmidrule(lr){5-7} \cmidrule(lr){8-10} \cmidrule(lr){11-13} 
& \multicolumn{2}{c}{Angular Error (°) $\downarrow$ } & \% Within $t^\circ$ $\uparrow$ 
& \multicolumn{2}{c}{Angular Error (°) $\downarrow$} & \% Within $t^\circ$ $\uparrow$ 
& \multicolumn{2}{c}{Angular Error (°) $\downarrow$} & \% Within $t^\circ$ $\uparrow$ 
& \multicolumn{2}{c}{Angular Error (°) $\downarrow$} & \% Within $t^\circ$ $\uparrow$  \\
\cmidrule(lr){2-3} \cmidrule(lr){4-4} \cmidrule(lr){5-6} \cmidrule(lr){7-7} \cmidrule(lr){8-9} \cmidrule(lr){10-10}  \cmidrule(lr){11-12} \cmidrule(lr){13-13}
& Mean & Median & 11.25° / 22.5° / 30° 
& Mean & Median & 11.25° / 22.5° / 30° 
& Mean & Median & 11.25° / 22.5° / 30° 
& Mean & Median & 11.25° / 22.5° / 30° \\
\midrule

PIFuHD \cite{saito2020pifuhd} &-&-&-&-&-&-&-&-&- & 22.39 & 19.26 & 23.0 / 60.1 / 77.0 \\
HDNet \cite{jafarian2021learning}  &-&-&-&-&-&-&-&-&- & 28.60 & 26.85 & 19.1 / 57.9 / 70.1 \\
ICON \cite{xiu2022icon}  &-&-&-&-&-&-&-&-&-  & 20.18 & 17.52 & 26.8 / 66.3 / 82.7 \\
ECON \cite{xiu2023econ} &-&-&-&-&-&-&-&-&-& 18.46 & 16.47 & 29.3 / 68.1 / 84.9 \\

Sapiens-0.3B & 
18.86 & 14.47 & 42.6 / 71.2 / 81.3 & 
\underline{12.54} & \underline{10.42} & \underline{56.2 / 88.0 / 94.6} &  
15.72 & 13.03 & 43.1 / 79.2 / 89.4 &
\underline{15.04} & \underline{12.22} & \underline{47.1 / 81.5 / 90.7}\\
Sapiens-2B &   
\textbf{16.04 }   &    \textbf{11.66}   &    \textbf{51.7 / 78.3 / 86.3}   &
\textbf{10.65}   &    \textbf{8.67 }  &   \textbf{65.5 / 92.5 / 96.7 }  &
\textbf{11.49 }  &     \textbf{9.07 } &   \textbf{62.3 / 90.2 / 95.4}  &
\textbf{12.14} & \textbf{9.62} & \textbf{60.2 / 89.1 / 94.7}\\
\midrule
Ours-Base & 17.33 & 12.36 & 47.7 / 75.9 / 84.5 & 14.10 & 11.32 & 50.3 / 83.9 / 91.8 & \underline{14.60} & 11.79 & 48.1 / \underline{82.3} / \underline{91.1} & 15.72 & 12.95 & 43.2 / 78.7 / 89.2\\
Ours-Large & \underline{17.15} & \underline{12.19} & \underline{48.4 / 76.3 / 84.7} & 13.96 & 11.23 & 50.7 / 84.2 / 92.1 & \underline{14.60} & \underline{11.66} & \underline{48.7} / 82.2 / 90.8 & 15.37 & 12.51 & 45.1 / 79.7 / 89.6\\

\bottomrule
\end{tabular}%
}
    \label{tab:normal_eval_all}
\end{table*}

\noindent\textbf{Soft Foreground Segmentation.}
\begin{table}
\centering
\caption{Cross dataset evaluation for soft foreground segmentation.} 
\scriptsize
\begin{tabular}{lccccc}
\toprule
\multirow{2}{*}{Method} & 
\multicolumn{3}{c}{PhotoMatte85} &
\multicolumn{2}{c}{PPM-100} \\
\cmidrule(lr){2-4} \cmidrule(lr){5-6}
& SAD $\downarrow$ & MSE $\downarrow$ & Conn $\downarrow$ & SAD $\downarrow$ & Conn $\downarrow$\\
\midrule
Zhong et al.~\tiny{\cite{zhong2024lightweight}}       & - & - & - & 90.28 & 84.09\\
BGMv2~\cite{lin2021real}  & - & - & - & 159.44 & 149.79\\
P3M-Net~\tiny{\cite{li2021privacy}} & 20.05 & 0.007 &  19.76 & 142.74 & 139.89\\
MODNet~\tiny{\cite{ke2022modnet}}    & 13.94 & 0.003 & 11.18 & 104.35 & 96.45\\
\midrule
Ours  &\textbf{ 5.85} & \textbf{0.0009 }& \textbf{5.60} & \textbf{78.17} & \textbf{74.72}\\
\bottomrule
\end{tabular}
\label{tab:matting_eval}

\end{table}
For human-centric dense prediction tasks, such as the ones we tackle in this work, we need to separate the foreground human from the background.
We do this by predicting the soft foreground mask using the Large version of our model.
The closest task to this is foreground matting\footnote{Our approach does not tackle the full matting problem, however, for the lack of better benchmark, we evaluate our approach on matting datasets.} for which we provide the results in \cref{tab:matting_eval}.
Our approach generalizes well, evident by the performance on PhotoMatte85 and PPM-100.
Note that some prior works listed in \cref{tab:matting_eval}, e.g., \citet{zhong2024lightweight}, are highly optimized for real-time portrait matting, making them considerably more efficient than ours.
We prioritize maintaining a unified architecture across without task-specific modifications, achieving superior accuracy and seamlessly integrating with the other two tasks.

\subsection{Ablation Studies}

\begin{table*}[t]
    \centering
    \scriptsize
    \setlength{\tabcolsep}{2.5pt}
    \caption{Ablation study: the effect of training data source, training data size, backbone size, and multi-task learning for depth estimation task evaluated on Goliath and Hi4D datasets.}
    \makebox[\textwidth][c]{
    \begin{minipage}[t]{0.31\textwidth}
        \centering
        \subcaption{\scriptsize Impact of training data.\label{tab:ablation:a}}
        \scalebox{0.9}{
        \begin{tabular}{l c c c c}
        \toprule
        \multirow{2}{*}{Source} & 
        \multicolumn{2}{c}{Goliath} & 
        \multicolumn{2}{c}{Hi4D} \\
        \cmidrule(lr){2-3} \cmidrule(lr){4-5} 
        & RMSE $\downarrow$ & AbsRel $\downarrow$ & RMSE $\downarrow$ & AbsRel $\downarrow$\\
        \midrule
        THuman2.0  &  0.495 & 0.017 & 0.137 & 0.040\\
        RenderPeople &  0.278 & 0.011 & 0.076 & 0.021\\
        Ours & 0.253 & 0.008 & 0.072 & 0.019 \\
        \bottomrule
        \end{tabular}
        }
    \end{minipage}
\hfill
    \begin{minipage}[t]{0.34\textwidth}
        \centering
        \subcaption{\scriptsize Impact of training data size.\label{tab:ablation:b}}
        \scalebox{0.9}{
        \begin{tabular}{l c c c c}
        \toprule
        \multirow{2}{*}{Dataset size} & 
        \multicolumn{2}{c}{Goliath} & 
        \multicolumn{2}{c}{Hi4D} \\
        \cmidrule(lr){2-3} \cmidrule(lr){4-5} 
        & RMSE $\downarrow$ & AbsRel $\downarrow$ & RMSE $\downarrow$ & AbsRel $\downarrow$ \\
        \midrule
        ViT-Base [60K]  & 0.324 & 0.011 & 0.101 & 0.028\\
        ViT-Base [150K] & 0.305 & 0.010 & 0.085 &  0.022 \\
        ViT-Base [300K] & 0.278 & 0.009 & 0.085 & 0.024\\
        \bottomrule
        \end{tabular}
        }
    \end{minipage}
\hfill
    \begin{minipage}[t]{0.3\textwidth}
        \centering
        \subcaption{\scriptsize Impact of model size.\label{tab:ablation:c}}
        \scalebox{0.9}{
        \begin{tabular}{l c c c c}
        \toprule
        \multirow{2}{*}{Arch.} & 
        \multicolumn{2}{c}{Goliath} & 
        \multicolumn{2}{c}{Hi4D} \\
        \cmidrule(lr){2-3} \cmidrule(lr){4-5} 
        & RMSE $\downarrow$ & AbsRel $\downarrow$ & RMSE $\downarrow$ & AbsRel $\downarrow$ \\
        \midrule
        ViT-Small  & 0.310 & 0.010 & 0.089 & 0.025  \\
        ViT-Base & 0.278 & 0.009 & 0.085 & 0.024\\
        ViT-Large & 0.253 & 0.008 & 0.072 & 0.019 \\
        \bottomrule
        \end{tabular}
        }
    \end{minipage}
}

    \vspace{5pt}
    \begin{minipage}{\textwidth}
        \centering
        \subcaption{\scriptsize Impact of multi-task training compared to single-task training.\label{tab:ablation:d}}
        \setlength{\tabcolsep}{5pt}
\resizebox{\linewidth}{!}{%
        \begin{tabular}{l c c c c c c c c c c c c }
        \toprule
        & 
        &
        \multicolumn{4}{c}{Depth} & 
    
        \multicolumn{4}{c}{Surface Normal} &
    
        \multicolumn{3}{c}{Matting} \\
        \multirow{2}{*}{} & &
        \multicolumn{2}{c}{Goliath} & 
        \multicolumn{2}{c}{Hi4D} &
        \multicolumn{2}{c}{Goliath} & 
        \multicolumn{2}{c}{Hi4D} &
        \multicolumn{1}{c}{PPM-100} &
        \multicolumn{2}{c}{PhotoMatte85}\\
        \cmidrule(lr){3-4} \cmidrule(lr){5-6} \cmidrule(lr){7-8} \cmidrule(lr){9-10} \cmidrule(lr){11-11} \cmidrule(lr){12-13}
        Setting & Params & RMSE $\downarrow$ & AbsRel $\downarrow$ & RMSE $\downarrow$ & AbsRel $\downarrow$ & MAE(°) $\downarrow$ & \% W $30^\circ$ $\uparrow$ &  MAE(°) $\downarrow$ & \% W $30^\circ$ $\uparrow$ & SAD $\downarrow$ & SAD $\downarrow$ & MSE $\downarrow$\\
        \midrule
          Single-task-Large & $3\times 0.34$B & 
         0.253 & 0.008 & 0.072 & 0.019 &
         15.24 & 89.19 & 
         15.37 & 89.56 &
         78.17 & 
         5.85 & 0.0009\\
         Single-task-Base & $3\times 0.12$B & 
         0.278 & 0.009 &
         0.085 & 0.024 &
         15.34 & 89.13 & 
         15.72 & 89.18 &
         90.86 & 
         7.97 & 0.0017\\
         Multi-task-Large & $1\times 0.35$B &
        0.270 & 0.009 &
         0.078 & 0.021 &
         15.27 & 89.12 &
         15.61 & 89.48 &
         66.08 & 
         5.40 & 0.0008 \\
        \bottomrule
        \end{tabular}%
}
    \end{minipage}
    \label{tab:ablation}
\end{table*}

In this section, we evaluate our design choices, including the impact of the synthetic data source, training data size, and model size. We also demonstrate that, since we use a single training dataset across multiple tasks to train a single model architecture, it is feasible to train a single model to perform all three tasks. We then demonstrate the accuracy of that model compared to three separately trained models specializing on each task. Unless otherwise stated, we use the depth estimation task for the ablation studies.

\noindent\textbf{Impact of data source.}
To compare the impact of the dataset quality, we render synthetic datasets from RenderPeople and THuman2.0 of similar size to our SynthHuman dataset and use these to train a Large depth estimation model with the same hyper-parameters as our model.
In \cref{tab:ablation:a} we see that the fidelity of the ground truth and the diversity of samples play a key role in achieving the best results. While the coarse depth estimated by each model is roughly the same, the model trained on SynthHuman is capable of capturing far more detail, shown in \cref{fig:rp_vs_thuman_vs_sx}.

\noindent\textbf{Impact of training data size.}
In \cref{tab:ablation:b}, we show the effect of the size of the training data. While even a small but high-fidelity dataset, as small as 60K, leads to reasonable accuracy for the relative depth estimation task, the model achieves better performance as we increase the training data size. This highlights that the diversity and fidelity of our dataset is considerable and the trainings do not saturate on a portion of dataset. Comparing this result with the last row of \cref{tab:ablation:c} also highlights that our synthetic data contributes positively as we scale on both the data and model size.

\noindent\textbf{Impact of model size.} 
Another aspect of training on relatively small datasets is interaction with model size. To ensure that our training data serves models of multiple sizes we train models with ViT variants of small, base, and large, with results reported in \cref{tab:ablation:c}. As expected, increasing model size lead to increase in the performance of the model.

\noindent\textbf{Multi-task model.}
Using a single dataset and a single model architecture allows us to easily train a single model with three convolution heads to perform multiple task learning. This is particularly important to combine soft foreground segmentation with depth and normal estimation, as for human-centric tasks it is needed to separate the human from the background. We observe that using three separate Large models yields slightly better results than a single Large multi-task model with one-third of the total parameters (3$\times$0.34B vs 0.35B), see \cref{tab:ablation:d}. 
Jointly training all three tasks in a multi-task model, however, performs better than three separate models with similar combined number of parameters (3$\times$0.12B for three Base models vs 0.35B for the Large multi-task one).

\begin{figure}
    \centering
    \scriptsize
    \begin{tabularx}{\linewidth}{CCCC}
          Input & Trained on & Trained on & Trained on our\\
          Image & \textbf{THuman2.1} & \textbf{RenderPeople} & \textbf{SynthHuman} \\
    \end{tabularx}
    \includegraphics[width=\linewidth]{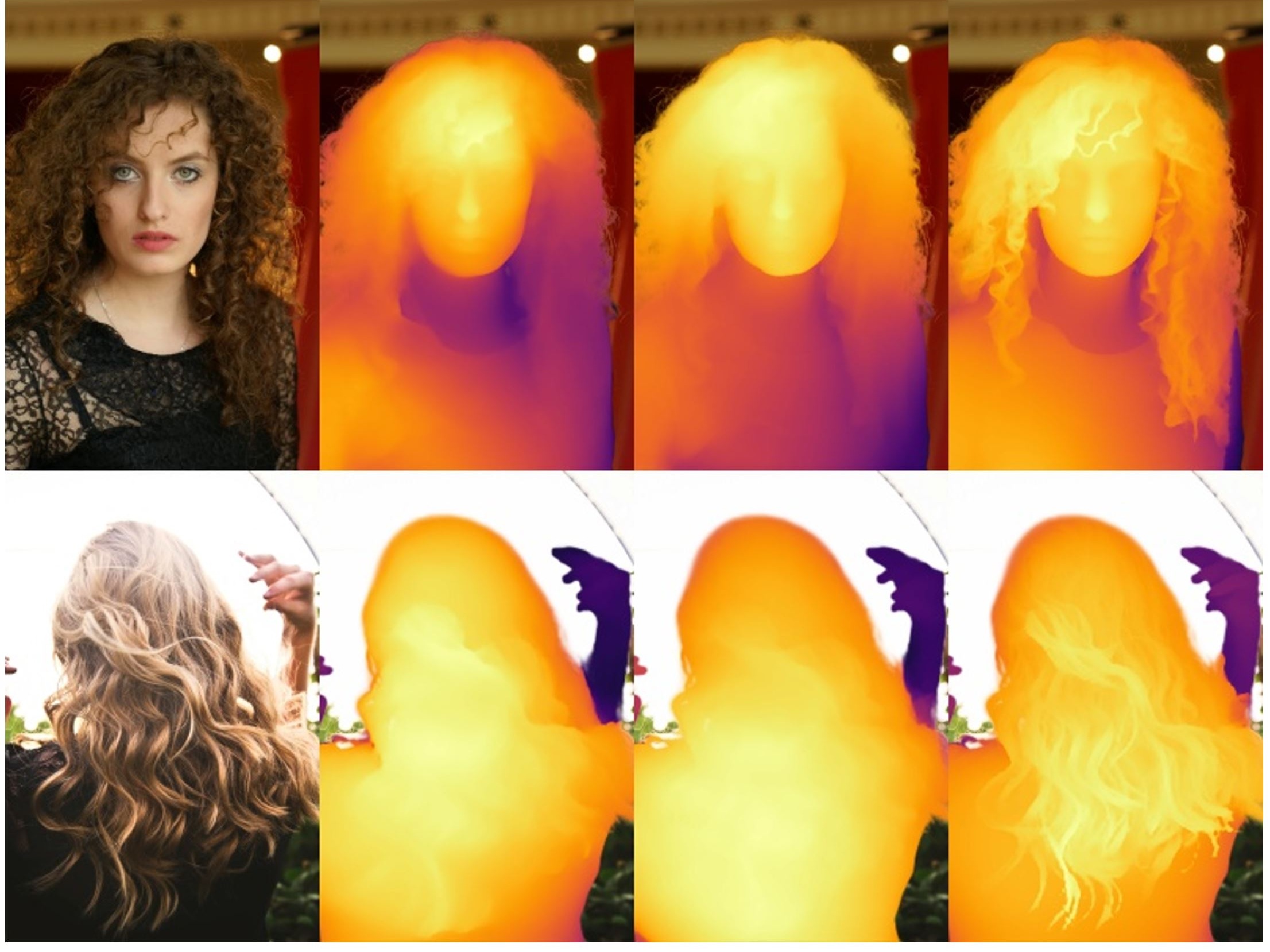}
    \vspace{-1.5em}
    \caption{Comparing the accuracy of models trained on Thuman2.1, RenderPeople and our SynthHuman dataset. Our dataset contains details (e.g., hair curls) that scan-based datasets struggle to capture. Note that the only change here is the training data (see \cref{tab:ablation:a}).}
    \label{fig:rp_vs_thuman_vs_sx}
\end{figure}

\section{Potential societal impact}
As for all human-centric computer vision, the models we train and demonstrate in this work could have lower accuracy for some demographic groups.
We find that our use of synthetic data helps in addressing any lack of fairness we discover in model evaluations, given the precise control we have over the training data distribution. 
Nevertheless, there are aspects of human diversity that are not yet represented by our datasets (see \cref{sec:limitations}), and there may also be lack of fairness that we have not yet discovered in evaluations.

A negative impact of the trend towards huge real datasets is difficulty in ensuring informed consent for training AI models, both from the rights holders and the people appearing in the images.
By demonstrating that models trained only on synthetic data can be as accurate as large foundational models, we hope to show that state-of-the-art human understanding need not be in tension with user privacy.

Another negative societal impact comes from the environmental cost of training and running inference on models that are larger than necessary.
By showing that human-centric vision models can achieve state-of-the-art accuracy at smaller model sizes, we hope to show that these techniques can be cost-effective and responsible in the use of compute resources, while sacrificing nothing in accuracy or robustness.

\section{Limitations}\label{sec:limitations}
Despite the strong generalizability of our trained models to real-world images, certain challenging scenarios still lead to failure cases, as illustrated in \cref{fig:failure}. 
For instance, extreme lighting conditions can introduce inaccuracies in defining surfaces. 
Our surface normal prediction model may misinterpret printed patterns on clothing or tattoos as distinct geometric structures instead of recognizing the underlying surface as continuous. 
Our relative depth estimation model struggles with rare scale variations. 
For example, when a baby is held in an adult’s hand, the model incorrectly perceives the large hand as significantly closer to the camera than the baby’s face.
Many of these failure cases could be mitigated by enhancing our synthetic dataset with more diverse assets and scene variations, thereby improving the model's robustness to such real-world diversity.

\begin{figure}
    \centering
    \includegraphics[width=\linewidth]{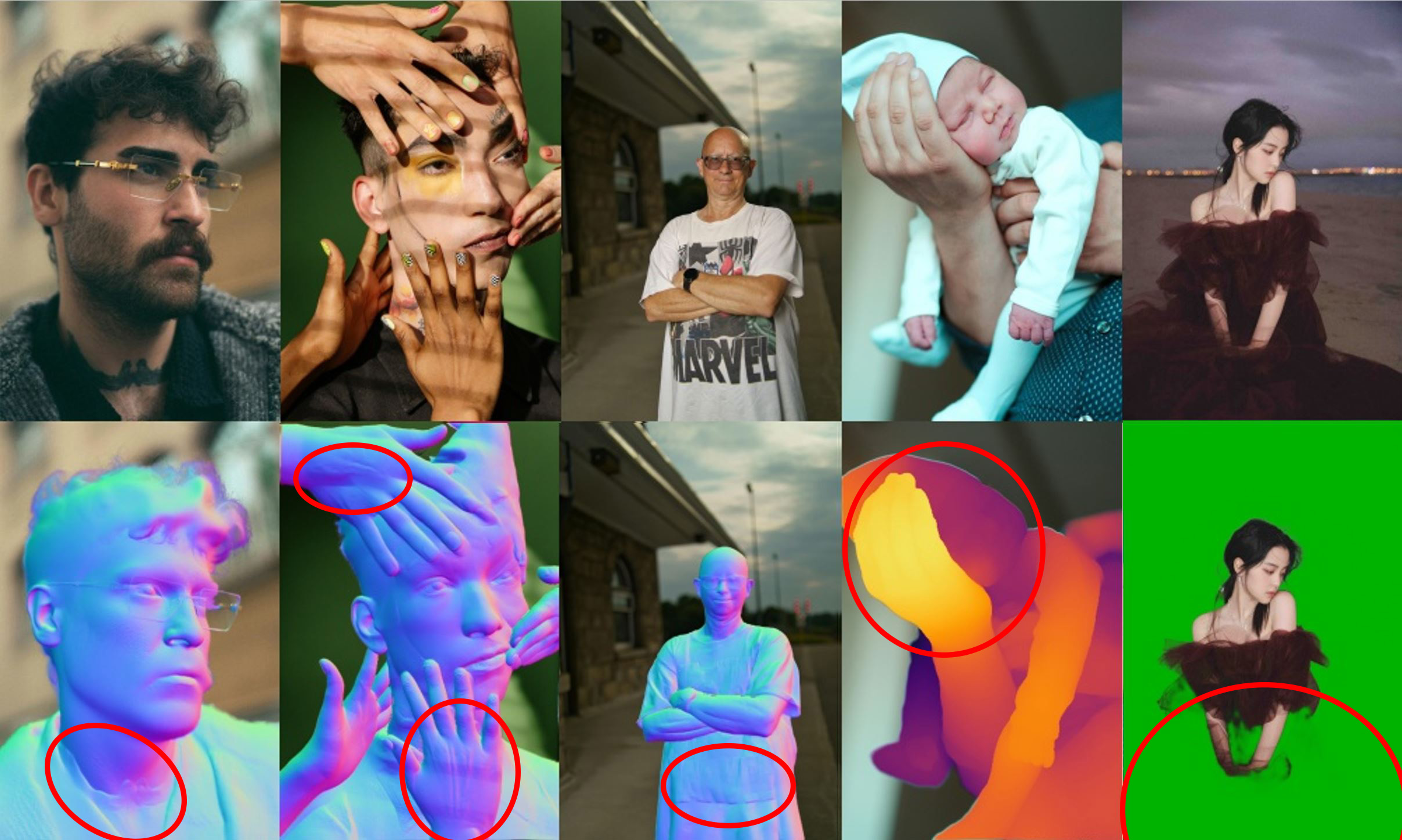}
    \vspace{-1.5em}
    \caption{Failures of our models in the presence of tattoos, extreme lighting, uncommon scale variations, and challenging clothing.
    }
    \label{fig:failure}
\end{figure}

\section{Conclusion}

We have demonstrated that it is possible to train accurate human-centric vision models without the need for large models, huge datasets, and complex methodologies.
This was achieved through procedural synthetic data that allows us to have both diverse and well annotated data.
Given the smaller dataset, we can train comparatively compact models in a fraction of the time (we can train $\sim$800 models with the compute used to train a single Sapiens-2B~\cite{khirodkar2024_sapiens} model), while achieving results that are on par with or surpasses existing state-of-the-art methods.
We release our datasets and models to encourage further research in this space.
\newpage
\section*{Appendix}
\appendix
In this supplementary material, we provide additional details on the data rendering and implementation of our method. We also provide additional qualitative and quantitative results. We encourage the readers to watch the supplementary video that contains additional results.

\section{Synthetic Data}
As described in the main paper, we use the data generation pipeline of \citet{hewitt2024look}, incorporating the updated face model of \citet{petikam2024eyelid}, to create SynthHuman. 
We extend this data generation pipeline for dense prediction tasks. 
Specifically, we make two main changes: re-defining the hair surface normals as well as re-defining the ground-truth depth and surface normals for transparent surfaces. Below, we delve into details of these changes.

Beyond these additional output streams, in SynthHuman we update the sampling procedure to increase the number unique identities and incorporate more diverse poses, lighting, and camera views. 
Specifically, we sample face/body shape (from training sources and a library of 3572 scans), expression and pose (from AMASS~\cite{mahmood2019amass}, MANO~\cite{romero2017embodied}, and more), texture (from high-res face scans with expression-based dynamic wrinkle maps blended in), hair (548 strand-level 3D hair, each with 100K+ strands), accessories (36 glasses, 57 headwear), 50 clothing tops, and environment (a mix of HDRIs and 3D environments).

\subsection{Hair Surface Normals}

In scan-based synthetic data, e.g., RenderPeople\cite{renderpeople}, ground-truth (GT) hair surface normals are obtained by renderings of scanned 3D human models. These scans represent hair with a coarse surface mesh.
In our synthetic data we explicitly represent hair as hundreds of thousands of individual 3D strands, enabling generation of GT depth, normals, alpha, etc. with strand-level granularity.
While dense strand-based 3D hair is a high-fidelity representation, when rendered from a portrait view they produce extremely high-frequency surface normals that appear noisy due to aliasing (See \cref{fig:hair_normals:a}).
For generating our ground-truth surface normals, we redefine our hair strand normals to align closer to the coarse hair mesh surface normals of THuman2.1~\cite{tao2021function4d} and Renderpeople~\cite{renderpeople}, in which the hair normals better represent the coarse shapes of hair clumps and volumes rather than individual strands.

We wish to generate hair surface normal images with the interpretablility of Sapiens~\cite{khirodkar2024_sapiens} hair normal training data, but without reducing the fidelity of our strand-based hair representation.
We first generate a voxel-grid volume with density based on the strand geometry that occupies the voxel. Using marching cubes we convert the volume to a coarse proxy mesh that approximates the combined hair strands (\cref{fig:hair_normals:b}) with interpretable normal vectors.
The proxy mesh does not capture fine-scale fly-away hair strand detail so we only use it to sample normal vectors.
For a point on a strand of our synthetic hair (head hair, facial hair, eyebrows, and eyelashes), we render the normal vector of the nearest proxy mesh surface which is smooth across the pixel grid, rather than the strand normals themselves which are noisy between pixels.
We render all hair strands this way to preserve the fidelity of our synthetic hair representation while generating normals representing the coarse shapes of the hair style (\cref{fig:hair_normals:c}).

\begin{figure*}
    \centering
    \footnotesize
    \begin{tabularx}{\linewidth}{CCC}
         \subcaption{Na\"{i}ve hair normals rendering \label{fig:hair_normals:a}}& 
         \subcaption{Generated hair proxy mesh\label{fig:hair_normals:b}} &
         \subcaption{Our hair normals transferred from proxy mesh\label{fig:hair_normals:c}}
    \end{tabularx}
    \vspace{-1em}\\
    \includegraphics[width=\linewidth]{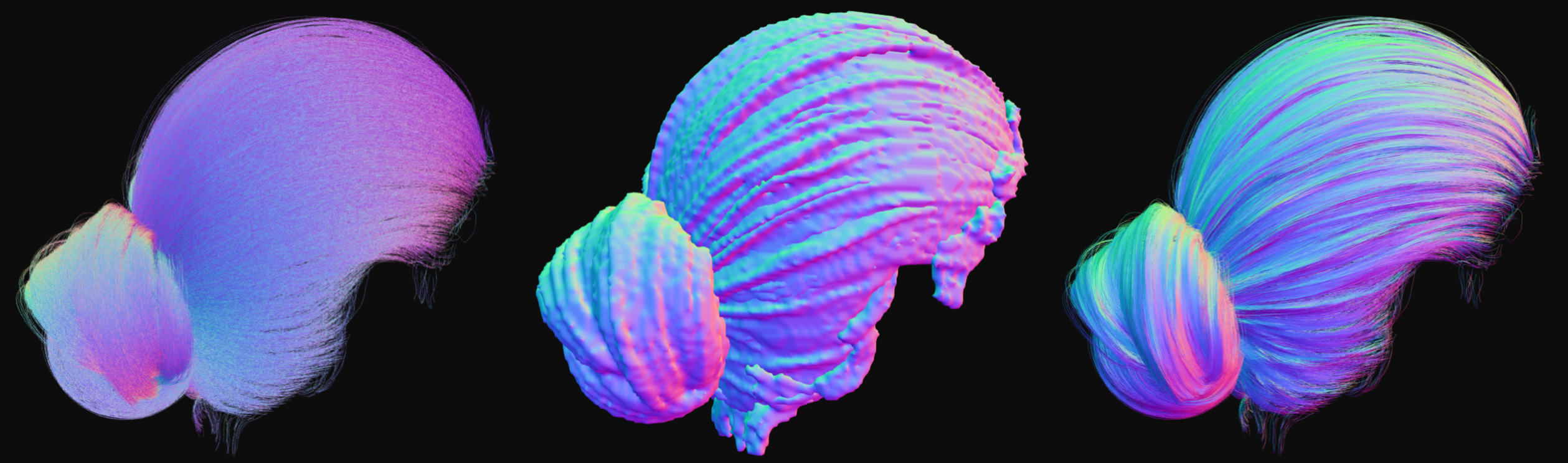}
    \caption{We generate interpretable strand-level synthetic hair normal GT training images by sampling normal directions from a proxy mesh representing the shape of the hair.}
    \label{fig:hair_normals}
\end{figure*}

\subsection{Ground-truth depth and normals of transparent surfaces.}

The predictions we show throughout this paper ignore the depth and normals of translucent surfaces like the lenses of glasses, instead predicting the depth and normals of the opaque surface visible behind the translucent media. For different applications we can control this behavior by choosing either to render the depth and normals of translucent surfaces or ignore them when generating our synthetic training images, as shown in \cref{fig:alpha_threshold}.

\begin{figure*}
    \centering
    \footnotesize
    \begin{tabularx}{\linewidth}{CCC}
         \subcaption{RGB training image \label{fig:alpha_threshold:a}}& 
         \subcaption{GT including translucent surfaces \label{fig:alpha_threshold:b}}& 
         \subcaption{GT ignoring translucent surfaces\label{fig:alpha_threshold:c}}
    \end{tabularx}
    \vspace{-1em}\\
    \includegraphics[width=\linewidth]{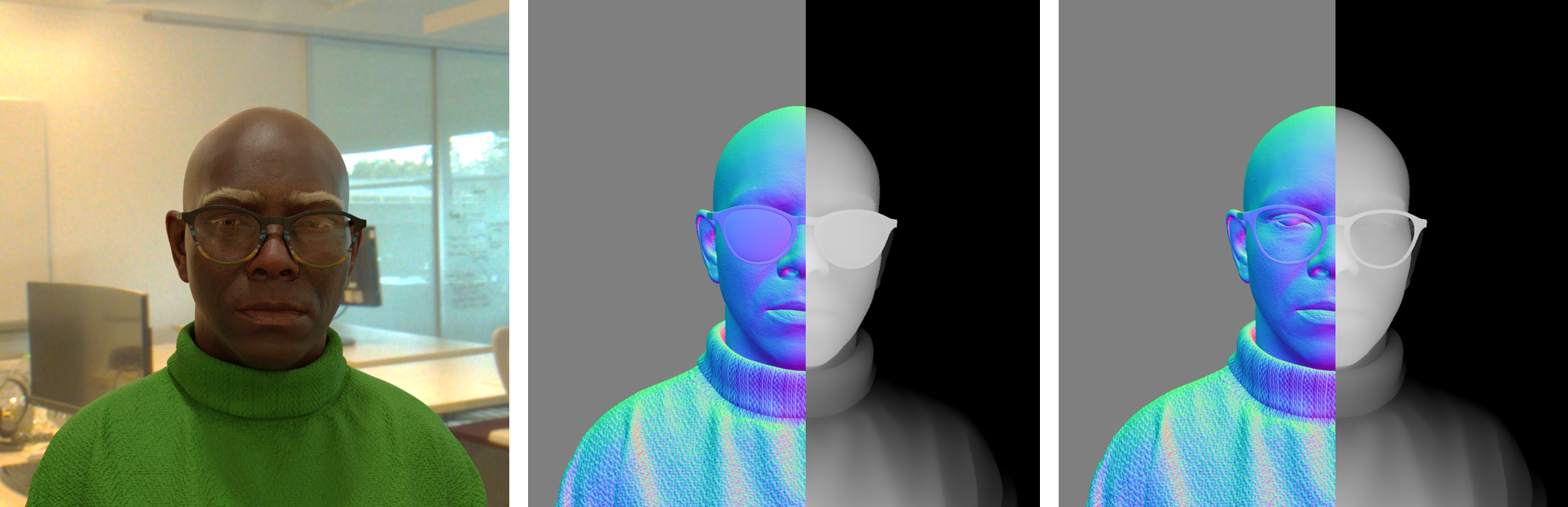}
    \caption{For different applications, we control how translucent surfaces are depicted in our generated normal and depth training images.}
    \label{fig:alpha_threshold}
\end{figure*}

\section{Experiments}

\subsection{Surface normal ground truth}

\begin{table*}
\centering
\scriptsize
\caption{Surface normal estimation using base model and blurring the output. Note that blurring results of our model leads to an increase in accuracy across all metrics, while blurring the output of Sapiens-0.3B makes little difference.}
\resizebox{\linewidth}{!}{%
\begin{tabular}{l cc c cc c cc c }
\toprule
\multirow{3}{*}{Method} & 
\multicolumn{3}{c}{Goliath-Face} & 
\multicolumn{3}{c}{Goliath-UpperBody} &
\multicolumn{3}{c}{Goliath-FullBody} \\
\cmidrule(lr){2-4} \cmidrule(lr){5-7} \cmidrule(lr){8-10}
& \multicolumn{2}{c}{Angular Error (°) $\downarrow$ } & \% Within $t^\circ$ $\uparrow$ 
& \multicolumn{2}{c}{Angular Error (°) $\downarrow$} & \% Within $t^\circ$ $\uparrow$ 
& \multicolumn{2}{c}{Angular Error (°) $\downarrow$} & \% Within $t^\circ$ $\uparrow$  \\
\cmidrule(lr){2-3} \cmidrule(lr){4-4} \cmidrule(lr){5-6} \cmidrule(lr){7-7} \cmidrule(lr){8-9} \cmidrule(lr){10-10} 
& Mean & Median & 11.25° / 22.5° / 30° 
& Mean & Median & 11.25° / 22.5° / 30° 
& Mean & Median & 11.25° / 22.5° / 30° \\
\midrule

Ours-Large &  17.15 & 12.19 & 48.4 / 76.3 / 84.7 & 13.96 & 11.23 & 50.7 / 84.2 / 92.1 & 14.60 & 11.66 & 48.7 / 82.2 / 90.8 \\

Ours with blur & \textbf{17.12} & \textbf{12.16} & \textbf{48.5 / 76.4 / 84.7} & \textbf{13.88} & \textbf{11.19} & \textbf{50.9 / 84.4 / 92.2} & \textbf{14.52} & \textbf{11.61} & \textbf{49.0 / 82.3 / 90.9} \\
\midrule
Sapiens-0.3B & 
18.86 & 14.47 & 42.6 / 71.2 / 81.3 & 
12.54 & 10.42 & 56.2 / 88.0 / 94.6 &  
15.72 & 13.03 & 43.1 / 79.2 / 89.4 \\

Sapiens-0.3B with blur & 
\textbf{18.84} & 14.47 & 42.6 / 71.2 / 81.3 & 
\textbf{12.51} & \textbf{10.40} & \textbf{56.3} / 88.0 / 94.6 &  
\textbf{15.69} & 13.03 & 43.1 / 79.2 / 89.4 \\

\bottomrule
\end{tabular}%
}
    \label{tab:normal_eval_all}
\end{table*}

Creating accurate surface normal annotations is very challenging for real data.
Most approaches rely on photogrammetry or reconstruction of relatively coarse surface meshes.
Both of the above approaches struggle with reconstructing thin or high frequency structures such as hair or folds in clothing.
They also struggle reconstructing the area around the eyes both due to thin structures (eyelashes), poor lighting due to self shadowing, and reflective surface of the eyeball.
This makes evaluating approaches that can capture such subtle details challenging as we may be seeing ceiling effect in results.

To demonstrate this we perform an experiment with taking the output of our surface normals models and blurring it using Gaussian Blur to reduce the fidelity of the output, rather than degrading the results this improves them on all metrics on the Goliath dataset.
This indicates that the ability to evaluate our models is hindered by quality of the annotations.

\subsection{Additional results for soft foreground segmentation.}

In \cref{tab:matting_p3m} we additionally show our soft foreground segmentation results on the two validation sets of the P3M dataset~\cite{li2021privacy}. While trained solely on synthetic data, our model achieves high accuracy on this challenging dataset. However, discrepancies arise due to differences in how the ground-truth alpha is obtained in our synthetic data compared to the P3M dataset, as well as variations in defining the most dominant human subjects in the scene, objects in hand, and other factors. This makes a fair comparison with methods trained on the P3M training set difficult.
To ensure a fair comparison, we conduct additional experiments. First, instead of training on SynthHuman, we train our model on P3M training subset. This shows that training on a dataset wherein ground-truth definitions match the test scenario is effective. In another experiment, we fine-tune our model, initially trained on SynthHuman, on the P3M training subset. By starting from a good initial weights (from our synthetic data), we show that fine-tuning on P3M and fixing the mismatches in the definition of foreground region is more effective, leading to the state-of-the-art results on most metrics.

\begin{table}[t]
\scriptsize
\centering
\caption{Evaluating soft foreground segmentation. Methods indicated by (*) are trained on the P3M training set.} 
\resizebox{\linewidth}{!}{%
\begin{tabular}{lcccccc}
\toprule
\multirow{2}{*}{Method} & \multicolumn{3}{c}{P3M-500-NP} & \multicolumn{3}{c}{P3M-500-P} \\
\cmidrule(lr){2-4} \cmidrule(lr){5-7}
 & SAD & SAD-T  & Conn & SAD & SAD-T  & Conn \\
\midrule
Zhong et al.*~\cite{zhong2024lightweight}       & 10.60 & \textbf{6.83} & 9.77 &  10.04 & \textbf{6.44} & 9.41 \\
BGMv2*~\cite{lin2021real} & 15.66 & 7.72  & 14.65 & 13.90 & 7.23 & 13.13 \\
P3M-Net*~\cite{li2021privacy} & 11.23 & 7.65  & 12.51 &  8.73 & 6.89 & 13.88 \\
MODNet~\cite{ke2022modnet}    & 20.20 & 12.48 & 18.41 & 30.08 & 12.22 & 28.61 \\
\midrule
Ours (trained on SynthHuman)  & 14.83 & 10.23 & 14.76 & 12.65 & 9.19 & 12.47 \\
Ours* (trained on P3M-train) & 12.30 & 9.46 & 12.14  & 11.48 & 8.29 & 11.35 \\
Ours* (trained on SynthHuman & \multirow{2}{*}{\textbf{9.12}} & \multirow{2}{*}{8.01} & \multirow{2}{*}{\textbf{8.94}} & \multirow{2}{*}{\textbf{8.05}} & \multirow{2}{*}{7.04} & \multirow{2}{*}{\textbf{7.90}}\\
{ } { } { } + by finetuned on P3M-train) & \\
\bottomrule
\end{tabular}%
}
\label{tab:matting_p3m}
\end{table}

\subsection{Additional results for depth estimation.}
\cref{tab:depth_eval_thuman} summarizes our results on the THuman2.1 dataset~\cite{tao2021function4d}. Following \cite{khirodkar2024_sapiens}, this synthetic dataset is rendered by placing THuman2.1 scans in HDRI environments. While we argue such synthetic data can act as a good resource for training, we do not consider them an ideal test benchmark. However, for completeness, we report our results on this dataset.
Following~\cite{khirodkar2024_sapiens}, we select 526 human scans from the THuman2.1 dataset and render 1,578 images to form our evaluation set. We observe that Sapiens~\cite{khirodkar2024_sapiens} achieves particularly strong results on this dataset, likely due to the close resemblance between THuman2.1 and RenderPeople which is used for their finetuning step. 
Our model, trained solely on SynthHuman dataset, also performs reasonably well on THuman2.1. 
However, we identify a significant difference between the quality of the rendered RGB images and depth ground-truth of THuman2.1 and those of SynthHuman. Particularly, as illustrated in Fig. 4 of the main paper, coarse and noisy scans of THuman2.1 lead to unrealistic RGB images and noisy ground-truth. 
To further analyze this, we utilize the remaining THuman2.1 scans to create a training set ($\sim$100k samples), rendered by placing a virtual camera around the scans placed in HDRI environments. Fine-tuning our depth model (initially trained on SynthHuman) on this additional data for only 25 epochs allows us to achieve on-par results with Sapiens. This shows that the difference in performance is primarily due to domain adaptation rather than inherent model capability.

\begin{table}
    \centering
    \footnotesize
    \caption{Evaluating depth estimation on THuman2.1 dataset. The results for Sapiens models indicated by (*) are re-evaluated on our rendered THuman2.1 evaluation subset, using exactly the same settings as in~\cite{khirodkar2024_sapiens}, except for the HDRIs, which may differ.} 
\resizebox{\linewidth}{!}{%
\begin{tabular}{lcccccccccc}
\toprule
\multirow{2}{*}{Method} & 
\multicolumn{3}{c}{TH2.0-Face} & 
\multicolumn{3}{c}{TH2.0-UprBody} & 
\multicolumn{3}{c}{TH2.0-FullBody} \\
\cmidrule(lr){2-4} \cmidrule(lr){5-7} \cmidrule(lr){8-10}
& RMSE & AbsRel & $\delta_1$ & RMSE & AbsRel & $\delta_1$ & RMSE & AbsRel & $\delta_1$ \\
\midrule
MiDaS-L~\cite{ranftl2020towards} & 0.114 & 0.097 & 0.925 & 0.398 & 0.271 & 0.868 & 0.701 & 0.689 & 0.782  \\
 MiDaS-Swin2~\cite{ranftl2020towards} & 0.050 & 0.036 & 0.995 & 0.122 & 0.081 & 0.948 & 0.292 & 0.171 & 0.862 \\
 DepthAny-B\cite{yang2025depth} & 0.039 & 0.026 & 0.999 & 0.048 & 0.028 & 0.999 & 0.061 & 0.030 & 0.999  \\
 DepthAny-L\cite{yang2025depth} & 0.039 & 0.027 & 0.999 & 0.048 & 0.027 & 0.999 & 0.060 & 0.030 & 0.999  \\
Sapiens-0.3B\cite{khirodkar2024_sapiens} &  0.012 & 0.008 & 1.000 & 0.015 & 0.009 & 1.000 & 0.021 & 0.010 & 1.000 \\
 Sapiens-2B~\cite{khirodkar2024_sapiens} &  0.008 & 0.005 & 1.000 &  0.010 & 0.006 & 1.000 & 0.016 & 0.008 & 1.000 \\
\midrule
Sapiens-0.3B* & 0.008 & 0.005 & 1.000 & 0.011 & 0.006 & 1.000 & 0.016 & 0.007 & 1.000 \\
Sapiens-2B* & 0.007 & 0.004 & 1.000 & 0.009 & 0.005 & 1.000 & 0.014 & 0.007 & 1.000 \\
\midrule
Ours (trained on SynthHuman)                  & 0.014 & 0.009 & 1.000 & 0.017 & 0.010 & 1.000 & 0.024 & 0.011 & 1.000  \\
Ours (trained on Thuman2.1) & 0.010 & 0.006 & 1.000 & 0.013 & 0.007 & 1.000 & 0.022 & 0.010 & 1.000\\

Ours (trained on SynthHuman & \multirow{2}{*}{0.008} & \multirow{2}{*}{ 0.005} & \multirow{2}{*}{1.000 } & \multirow{2}{*}{0.012} & \multirow{2}{*}{ 0.006 } & \multirow{2}{*}{1.000} & \multirow{2}{*}{0.018} & \multirow{2}{*}{0.008} & \multirow{2}{*}{1.000}\\
{ } { } { } + by finetuned on Thuman2.1) & \\
\bottomrule
\end{tabular}%
}
    \label{tab:depth_eval_thuman}
\end{table}

\begin{figure}
    \centering
    \tiny
    \begin{tabularx}{\linewidth}{CCCCC}
          Input & Normals at& Depth at& Normals at & Depth at \\
          Image & {\tiny $1024\times 1024$} & {\tiny $1024\times 1024$} & {\tiny $512\times 512$} & {\tiny $512\times 512$}\\
    \end{tabularx}
    \includegraphics[width=\linewidth]{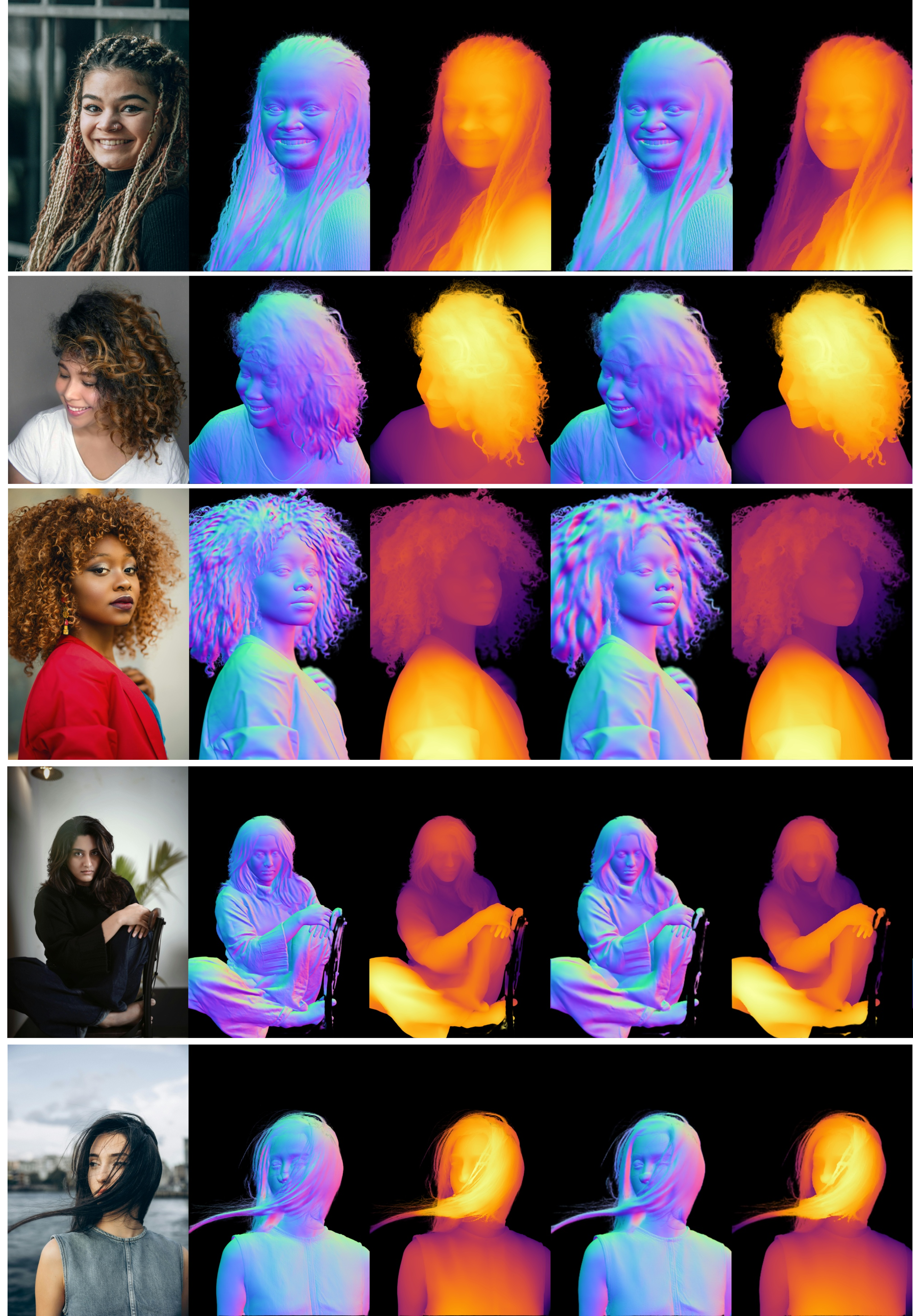}
    \caption{The Resizer module allows us to use arbitrary input size at test time. Higher resolution input provides more details to the model, thus it can capture more details in the depth and surface normals predictions.}
    \label{fig:high_res}
\end{figure}

\subsection{Remark on Resizer.}
In our method, we use the Resizer module to handle any resolution while running the ViT encoder on the fixed-size version of the image ($384\times 384$). While we use the resolution of $512\times 512$ (with 512 pixels being the height of SynthHuman images) for all the experiments in this paper, Resizer module allows us to make predictions at higher resolution. In Fig.~\ref{fig:high_res}, we show the output of the model when tested with input images of size $512\times 512$ versus $1024\times 1024$ (after padding to make square, if needed). We noticed that while still performing very fast, larger input resolution provides the model with far more details for all tasks.

\begin{figure*}
    \centering
    \includegraphics[width=\linewidth]{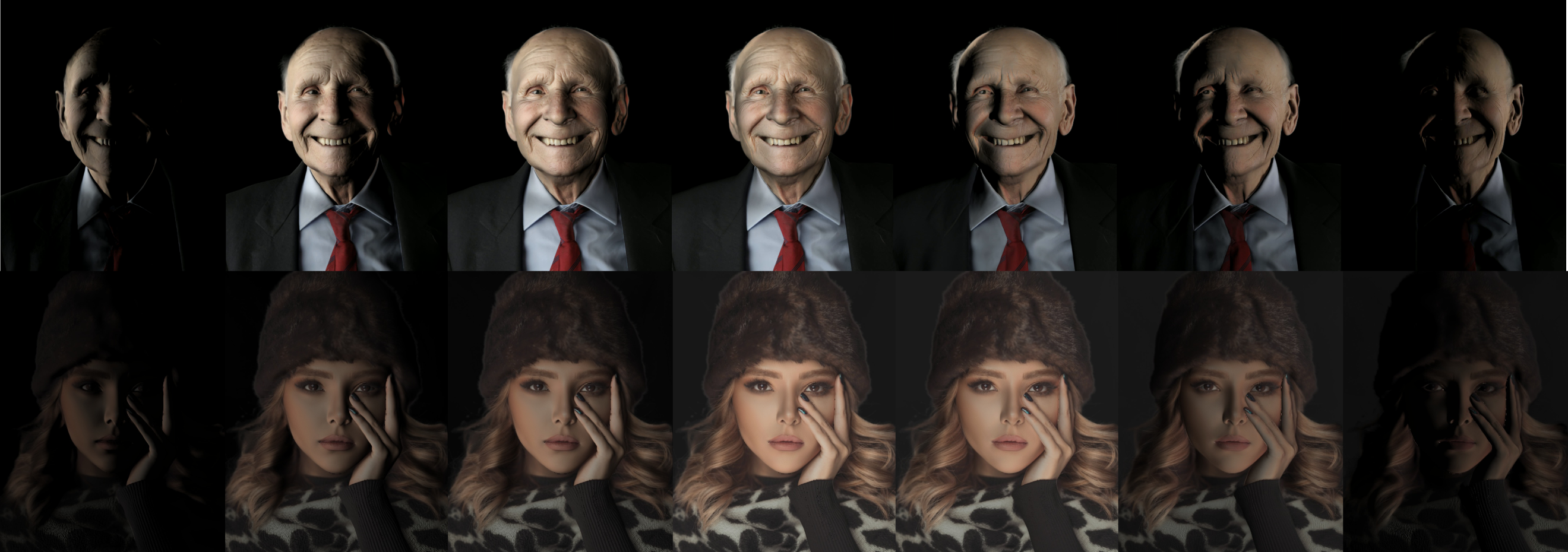}
    \caption{Examples of simple relighting using surface normals predicted by our model on in-the-wild data.}
    \label{fig:relighting}
\end{figure*}
\begin{figure*}
    \centering
    \includegraphics[width=\linewidth]{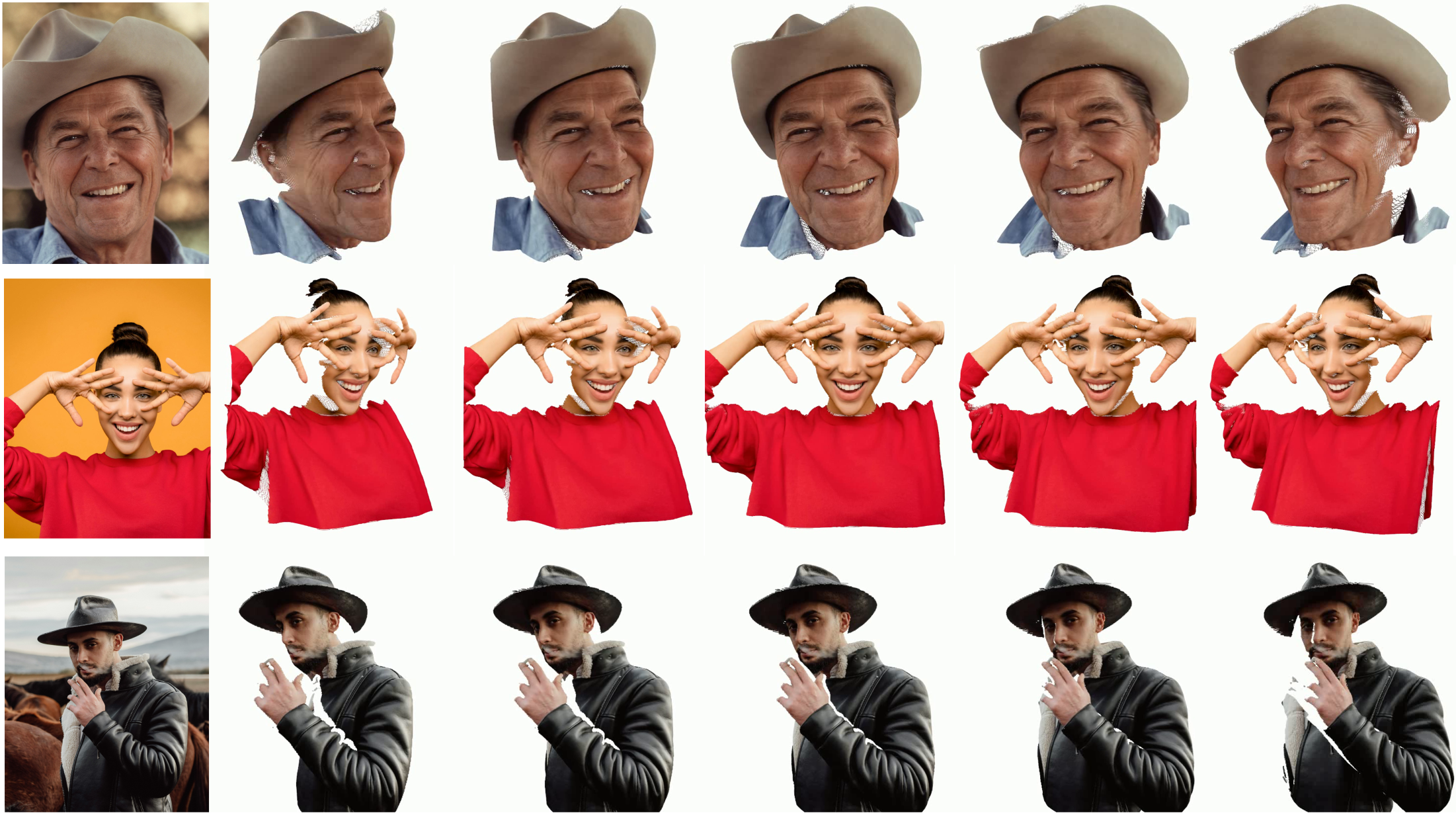}
    \caption{Results of our depth prediction model on in-the-wild images rendered as a point cloud from different viewpoints.}
    \label{fig:depth_3d}
\end{figure*}
\begin{figure*}
    \centering
    \includegraphics[width=\linewidth]{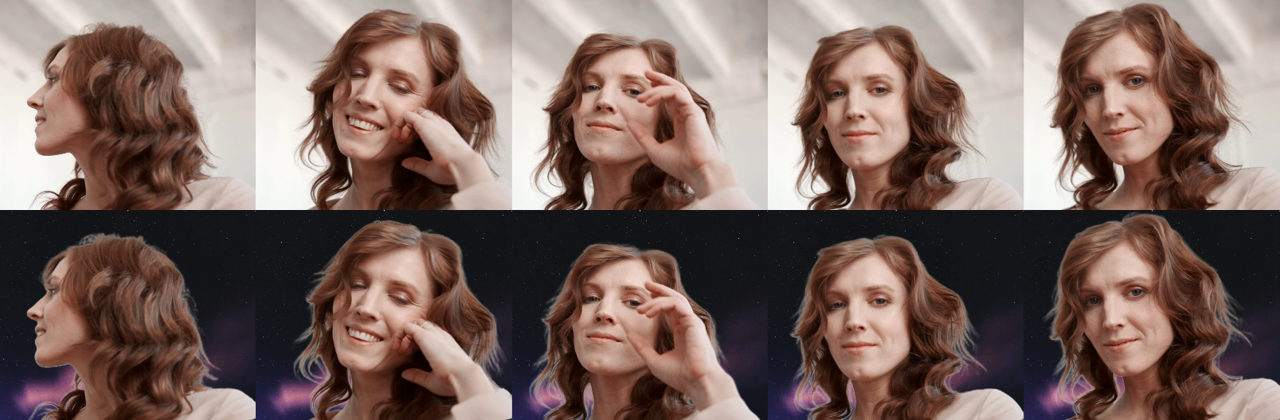}
    \caption{Background replacement demonstrated using results from our matting model on in-the-wild images.}
    \label{fig:fgbgseg}
\end{figure*}

\subsection{Applications of Dense Prediction Tasks}
In this section, we provide potential downstream applications for the dense prediction tasks we addressed in this paper. Particularly, we use our surface normal estimation model for a simple relighting. We demonstrate how we can use our depth estimation model to generate a 2.5D representation from a single image. And finally, we show that our soft foreground prediction model can be used for background replacement (e.g., in video conferencing).

\paragraph{Simple relighting from Normals.}
As a potential downstream application, we use our normal estimation model in a relighting pipeline to re-render images under novel lighting conditions. To this end, we first predict a surface normal map for an input image. This predicted normals, which capture fine geometric details, serve as the foundation for our relighting process. For a given image, we compute per-pixel shading based on a Lambertian reflectance model where the intensity is modulated by the cosine of the angle between each predicted normal and an externally specified light direction. To further enhance realism, we incorporate an ambient term, ensuring that areas not directly illuminated still receive a baseline level of light. As illustrated in Fig.~\ref{fig:relighting}, this re-rendering approach produces a visually plausible approximation of how the scene would appear under different lighting conditions.

\paragraph{2.5D representation from depth.}
We further demonstrate that our relative depth estimation model is capable of estimating the 2.5D representation of a given image. For a given image, the estimated depth map is then unnormalized using a reasonable guess of a range, which we use to generate a 3D point cloud of the visible scene. By rendering this point cloud from multiple novel viewpoints, as illustrated in Fig.~\ref{fig:depth_3d}, we demonstrate that our model captures challenging depth relations with remarkable fidelity. For example, the reconstructed geometry preserves correct facial proportions, clearly positions a hand in front of the body, and accurately depicts the shape of a hat on the head. These results illustrate that our relative depth model reliably encodes fine-grained depth cues, enabling effective 2.5D reconstruction from a single image.

\paragraph{Background replacement from segmentation.}
In addition to its primary role in supporting dense prediction tasks, our soft foreground segmentation model serves as a robust standalone solution for applications that require precise subject extraction. For example, as shown in Fig.~\ref{fig:fgbgseg}, our approach enables reliable background replacement, which is particularly valuable for video conferencing. By accurately separating the human subject and preserving fine details such as hair strands, our model ensures high-quality background substitution, demonstrating its effectiveness in real-world scenarios.

\subsection{Implementation Details}

During training, we apply various augmentations to enhance model robustness. For geometric transformations, we use random scaling to simulate zooming in or out of the image and its corresponding ground truth. Additionally, random shift augmentation is applied to simulate the shifting of ROI in both the image and GT.
For appearance augmentations, we apply random blurring to the image, with the blur strength proportional to the image size, simulating lenses with poor modulation transfer function (MTF).We adjust image brightness by adding a constant offset within a specified range and adjust the contrast using the formula: $$\text{img} = (\text{img} - 0.5) × (1 + \text{contrast}) + 0.5$$
Additionally, we randomly alter the hue and saturation, apply JPEG compression, and occasionally convert the image from BGR to greyscale. These appearance augmentations are applied with a specified probability. Following \citet{hewitt2024look}, we also introduce random ISO noise, inspired by real camera noise, to enhance training. This noise is a combination of image intensity-dependent Poissonian noise and intensity-independent Gaussian noise.

\subsection{Goliath Test Set}
\cref{tab:goliath_details} gives the frame and camera indices which are used for selecting and rendering ground truth for the evaluation set used in our work. We render the normal and depth images at $667\times1024$ resolution using Blender.
\newcolumntype{Q}[1]{>{\hsize=#1\hsize\centering\arraybackslash}X}

\begin{table*}
    \centering
    \footnotesize
    \begin{tabularx}{\linewidth}{Q{0.075} X Q{0.11} X}
\toprule
    Subset & Camera IDs & Subject & Frame IDs \\
    \midrule
    \multirow{8}{=}{Face} & \multirow{8}{=}{\emph{401650}, 401645, 401655, 401894, \emph{401962}, 402601, 402792, 402807, 402871, 402875, 402980, 403072} & AXE977 & 02858, 13148, 23438, 28085, 29114, 34733, 49044, 62745, 75355, 87055, 99319, 110328, 121299, 132449, 139288, 140317\\
    & & QZX685 & 03339, 13089, 22839, 28404, 29379, 30354, 46874, 62806, 74953, 85733, 97027, 107481, 119069, 131633, 132608, 133583\\
    & & XKT970 & 03178, 12868, 22558, 28225, 29194, 30163, 37300, 53338, 66207, 77489, 88184, 98424, 108787, 119398, 124264, 125233\\
    & & QVC422 & 03280, 13990, 24730, 28636, 29707, 30778, 31849, 33856, 52762, 69555, 82046, 93762, 105020, 116706, 123621, 124692\\
    \midrule
    \multirow{8}{=}{Upper Body} & \multirow{8}{=}{401541, 400874, 400883, 400894, 400895, 400898, 400926, 400929, 400933, 400934, 400936, 401534} & AXE977 & 00202, 02944, 05686, 08428, 11170, 13261, 14175, 22719, 25761, 28654, 31695, 34739, 37780, 40673, 43714, 46757\\
    & & QZX685 & 00227, 02981, 05735, 08489, 11243, 13544, 14462, 22813, 25868, 28773, 31825, 34881, 37935, 40838, 43890, 46944\\
    & & XKT970 & 00313, 03049, 05785, 08521, 11257, 13358, 14270, 22906, 25941, 28827, 31863, 34900, 37936, 40822, 43857, 46892\\
    & & QVC422 & 00207, 02913, 05619, 08325, 11031, 13150, 14052, 22493, 25498, 28354, 31362, 34368, 37373, 40229, 43236, 46242\\
    \midrule
    \multirow{8}{=}{Full Body} & \multirow{8}{=}{401156, 401150, 401185, 401191, 402359, 402401, 402432, 402435, 402547, 402551, 402636, 402689} & AXE977 & 00202, 02944, 05686, 08428, 11170, 13261, 14175, 22719, 25761, 28654, 31695, 34739, 37780, 40673, 43714, 46757 \\
    & & QZX685 & 00227, 02981, 05735, 08489, 11243, 13544, 14462, 22813, 25868, 28773, 31825, 34881, 37935, 40838, 43890, 46944\\
    & & XKT970 & 00313, 03049, 05785, 08521, 11257, 13358, 14270, 22906, 25941, 28827, 31863, 34900, 37936, 40822, 43857, 46892\\
    & & QVC422 & 00207, 02913, 05619, 08325, 11031, 13150, 14052, 22493, 25498, 28354, 31362, 34368, 37373, 40229, 43236, 46242\\
\bottomrule
\end{tabularx}
    \caption{Goliath evaluation set camera and frame selection. There are 12 cameras per subset and 16 frames per camera. Note 401650 is missing for calibration for subject XKT970 and 401962 is missing calibration for subject QZX685, so in total there are 2272 images.}
    \label{tab:goliath_details}
\end{table*}

\subsection{Additional Qualitative Results}
In this section, we provide additional qualitative results of our approach and compare them with Sapiens-2B models in Fig.~\ref{fig:additional_results}.

\begin{figure*}
\centering
    \scriptsize
    \begin{tabularx}{0.9\textwidth}{CCCCCCCCCC}
         Input & \multicolumn{2}{c}{Ours} &  \multicolumn{2}{c}{Sapiens-2B~\cite{khirodkar2024_sapiens}} & Input & \multicolumn{2}{c}{Ours} & \multicolumn{2}{c}{Sapiens-2B~\cite{khirodkar2024_sapiens}} \\
          Image & Normals & Depth & Normals & Depth & Image & Normals & Depth & Normals & Depth\\
    \end{tabularx}
    \includegraphics[width=.9\textwidth]{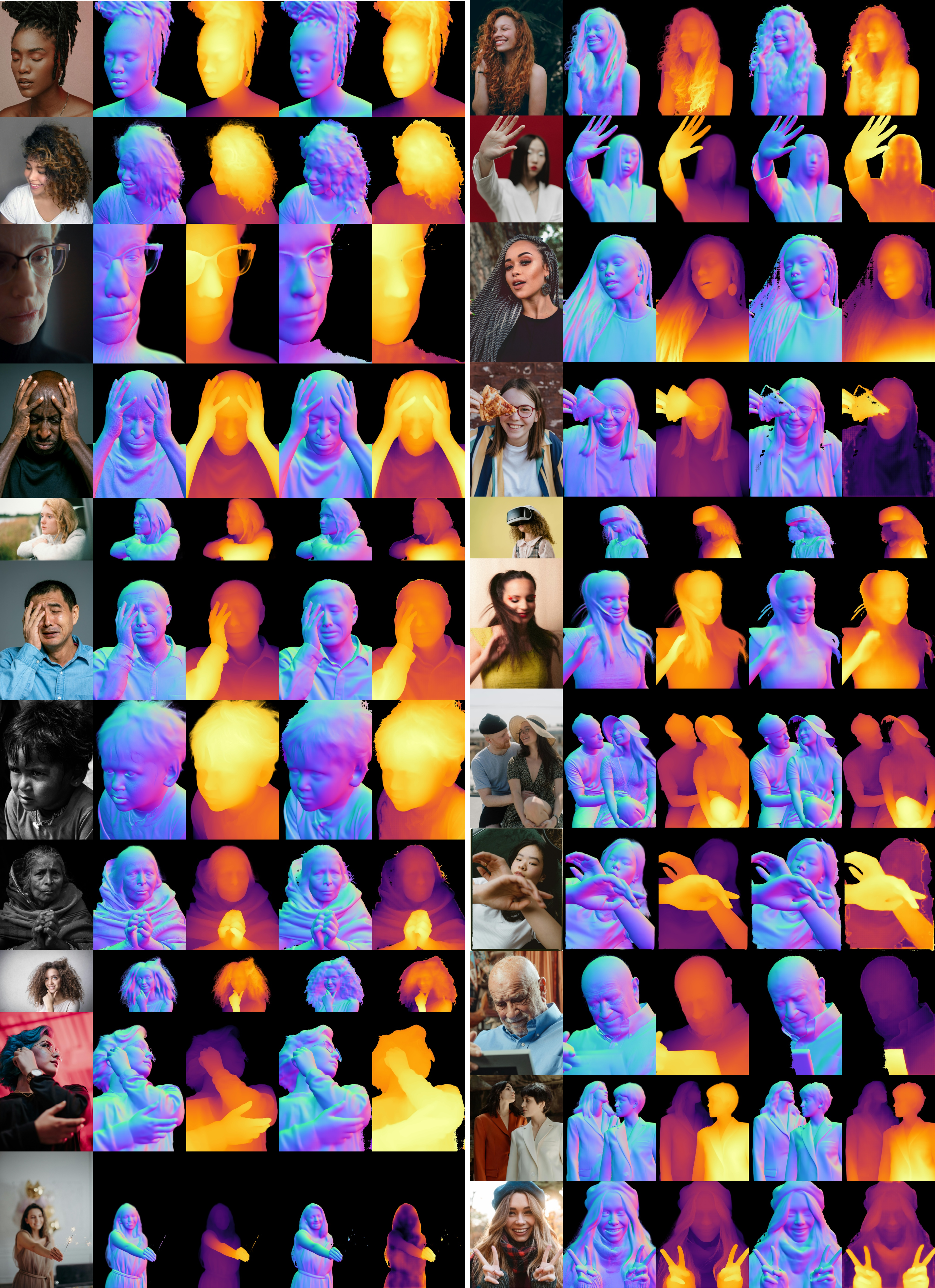}
    \caption{Additional qualitative comparisons.}
    \label{fig:additional_results}
\end{figure*}

{\small
\bibliographystyle{ieeenat_fullname}
\bibliography{egbib}
}

\end{document}